\documentclass[10pt,twocolumn,letterpaper]{article}

\usepackage[accsupp]{axessibility}
\usepackage{iccv}   
\usepackage{multirow}
\usepackage{pifont}
\definecolor{iccvblue}{rgb}{0.21,0.49,0.74}
\usepackage[pagebackref,breaklinks,colorlinks,allcolors=iccvblue]{hyperref}

\title{OmniVTON: Training-Free Universal Virtual Try-On}

\author{
Zhaotong Yang{$^{1}$}, Yuhui Li{$^{1}$}, Shengfeng He{$^{2}$}, Xinzhe Li{$^{1}$}, Yangyang Xu{$^{3}$}, Junyu Dong{$^{1}$}, Yong Du{$^{1}$}\thanks{Corresponding author (csyongdu@ouc.edu.cn).}\\
{$^{1}$}Ocean University of China, \\
{$^{2}$}Singapore Management University, {$^{3}$}Harbin Institute of Technology (Shenzhen)\\
}

\begin{document}
\teaser{
    \centering
    \includegraphics[width=0.95\textwidth]{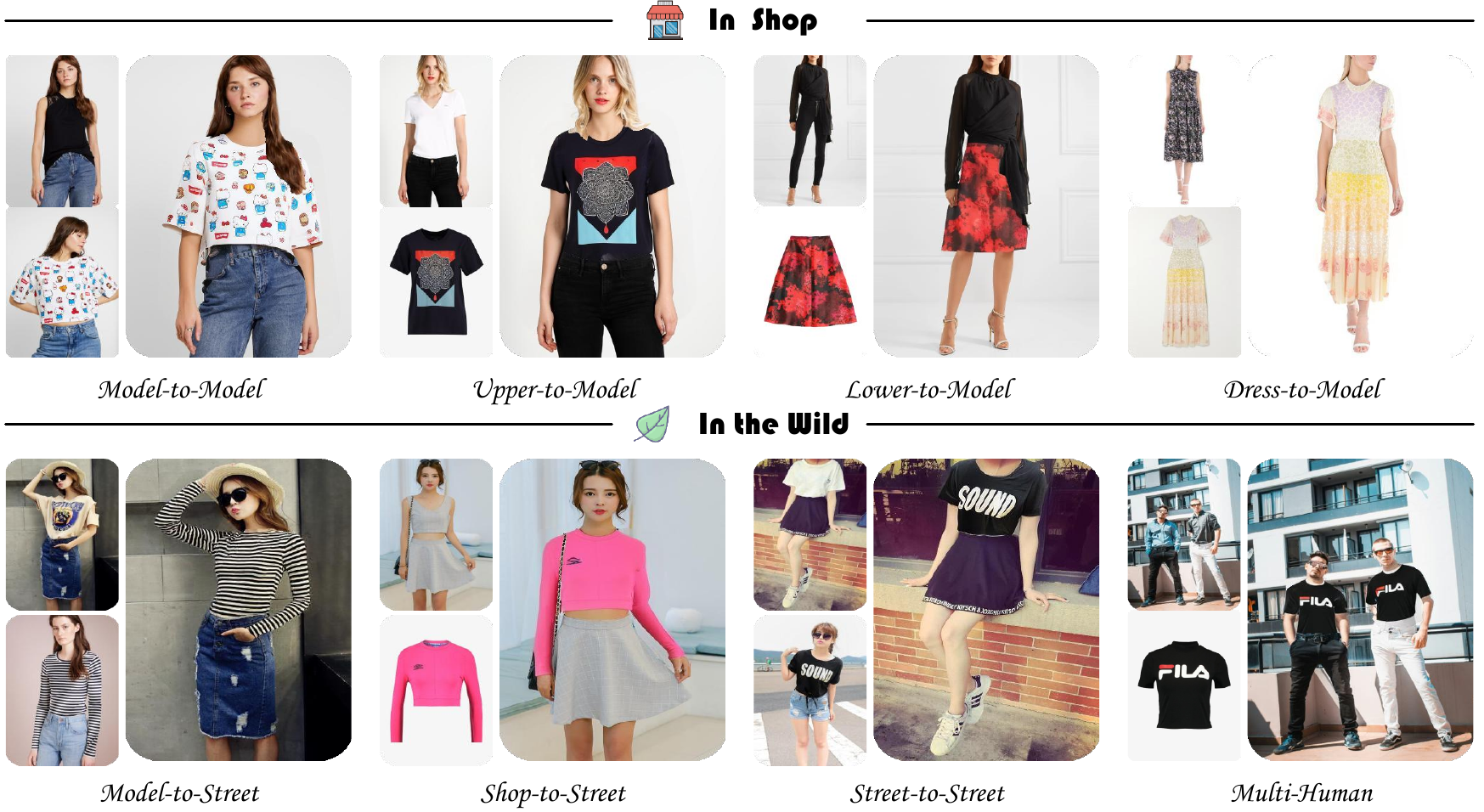}  
    \vspace{-2mm}\caption{We propose OmniVTON, a training-free universal virtual try-on framework that unifies both in-shop and in-the-wild scenarios while preserving garment details and ensuring pose consistency.}
    \label{fig:teaser}    
    \vspace{-4mm}
}
\maketitle

\begin{abstract}
Image-based Virtual Try-On (VTON) techniques rely on either supervised in-shop approaches, which ensure high fidelity but struggle with cross-domain generalization, or unsupervised in-the-wild methods, which improve adaptability but remain constrained by data biases and limited universality. A unified, training-free solution that works across both scenarios remains an open challenge. We propose OmniVTON, the first training-free universal VTON framework that decouples garment and pose conditioning to achieve both texture fidelity and pose consistency across diverse settings. To preserve garment details, we introduce a garment prior generation mechanism that aligns clothing with the body, followed by continuous boundary stitching technique to achieve fine-grained texture retention. For precise pose alignment, we utilize DDIM inversion to capture structural cues while suppressing texture interference, ensuring accurate body alignment independent of the original image textures. By disentangling garment and pose constraints, OmniVTON eliminates the bias inherent in diffusion models when handling multiple conditions simultaneously. Experimental results demonstrate that OmniVTON achieves superior performance across diverse datasets, garment types, and application scenarios. Notably, it is the first framework capable of multi-human VTON, enabling realistic garment transfer across multiple individuals in a single scene. Code is available at \url{https://github.com/Jerome-Young/OmniVTON}.
\end{abstract}

\section{Introduction}
\label{sec:intro}
Image-based virtual try-on (VTON) transforms online shopping by seamlessly integrating garment images with target human bodies to generate natural-looking results that conform to body poses while preserving texture consistency. It enhances the shopping experience by reducing uncertainty and minimizing return rates.

Existing VTON methods are designed for either in-shop or in-the-wild scenarios. Supervised approaches~\cite{hr-vton,vitonhd,dci-vton,d4-vton,ootd} dominate in-shop settings, achieving high-fidelity synthesis using paired training data but struggling with cross-domain/-scenario generalization. Conversely, in-the-wild methods~\cite{street-tryon} leverage unsupervised learning to improve adaptability across diverse input sources (\eg, Shop-to-Street, Model-to-Model, \etc) but remain constrained by data distribution biases and limited universality. Both paradigms rely on dedicated models trained for specific conditions, making large-scale dataset construction across all garment categories, styles, and human poses highly impractical. This fragmentation underscores the need for a unified VTON framework that can generalize across domains without requiring additional training.

To enable a training-free VTON framework, two critical challenges must be addressed:

\noindent i) \textit{Fine-grained Texture Consistency}: Without a dedicated training phase, it is difficult to establish garment-body alignment while preserving intricate texture details. Conventional methods rely on learned deformation priors, which are unavailable in a training-free setting.

\noindent ii) \textit{Human Pose Alignment}: Existing methods condition on keypoints~\cite{openpose} or DensePose maps~\cite{densepose}, requiring retraining for cross-modal feature fusion. Without explicit pose supervision, training-free approaches would struggle with pose consistency, especially for garments with ambiguous structures like sleeveless vests.

To tackle these issues, we propose \textit{OmniVTON}, a training-free virtual try-on framework that leverages off-the-shelf diffusion models through a progressive garment adaptation mechanism. For texture preservation, we introduce Structured Garment Morphing (SGM), which ensures seamless garment-body integration while maintaining fine-grained texture details. First, a pseudo-person image is generated via semantic-guided garment completion. Then, multi-part semantic correspondence between the pseudo-person and the source person is established using a predicted segmentation map and skeleton data. Finally, localized transformations dynamically adjust different garment regions to achieve an anatomically accurate alignment, producing a structurally coherent adaptation result.
To address inconsistencies along garment boundaries, we propose Continuous Boundary Stitching (CBS), which refines the transitions between segmented regions to ensure seamless integration. By leveraging semantic interactions between the latent features of the original garment image and the garment-infused image, CBS eliminates harsh edges and discontinuities, preserving the visual realism of the final synthesis.

For pose information injection, a naive approach is to directly apply DDIM Inversion~\cite{ddim}, which preserves structural information by replacing the initial random noise with inversion noise from the source person. However, this also introduces unwanted texture contamination. To address this, we propose Spectral Pose Injection (SPI), which selectively integrates pose cues while suppressing texture interference. By leveraging spectral analysis in the latent space, SPI retains low-frequency inversion noise for structural consistency while replacing high-frequency components with random noise to enhance generative flexibility. This frequency-aware modulation maintains pose fidelity while preventing residual texture artifacts.

Comprehensive experiments demonstrate that OmniVTON surpasses existing methods across multiple benchmarks, both qualitatively and quantitatively, producing high-fidelity try-on results while offering new insights into virtual try-on. Additionally, it showcases strong generalizability across various scenarios, datasets, and clothing types. The key contributions of this work include:
\begin{itemize}
\item We propose OmniVTON, a training-free universal VTON framework that unifies in-shop and in-the-wild scenarios, significantly advancing the state of the art.
\item We introduce Structured Garment Morphing, ensuring fine-grained texture preservation and seamless garment adaptation across diverse clothing types and scenarios.
\item We develop Spectral Pose Injection and Continuous Boundary Stitching to effectively integrate pose information and refine textures, producing pose-consistent and texture-coherent try-on results.
\item Our method achieves state-of-the-art results across multiple evaluation metrics, demonstrating superior quality, generalizability, and scalability. Notably, it is the first to enable multi-human VTON, facilitating realistic garment transfer across multiple individuals.
\end{itemize}

\section{Related work}
\label{sec:rela}
\begin{figure*}
	\centering
	\includegraphics[width=\linewidth]{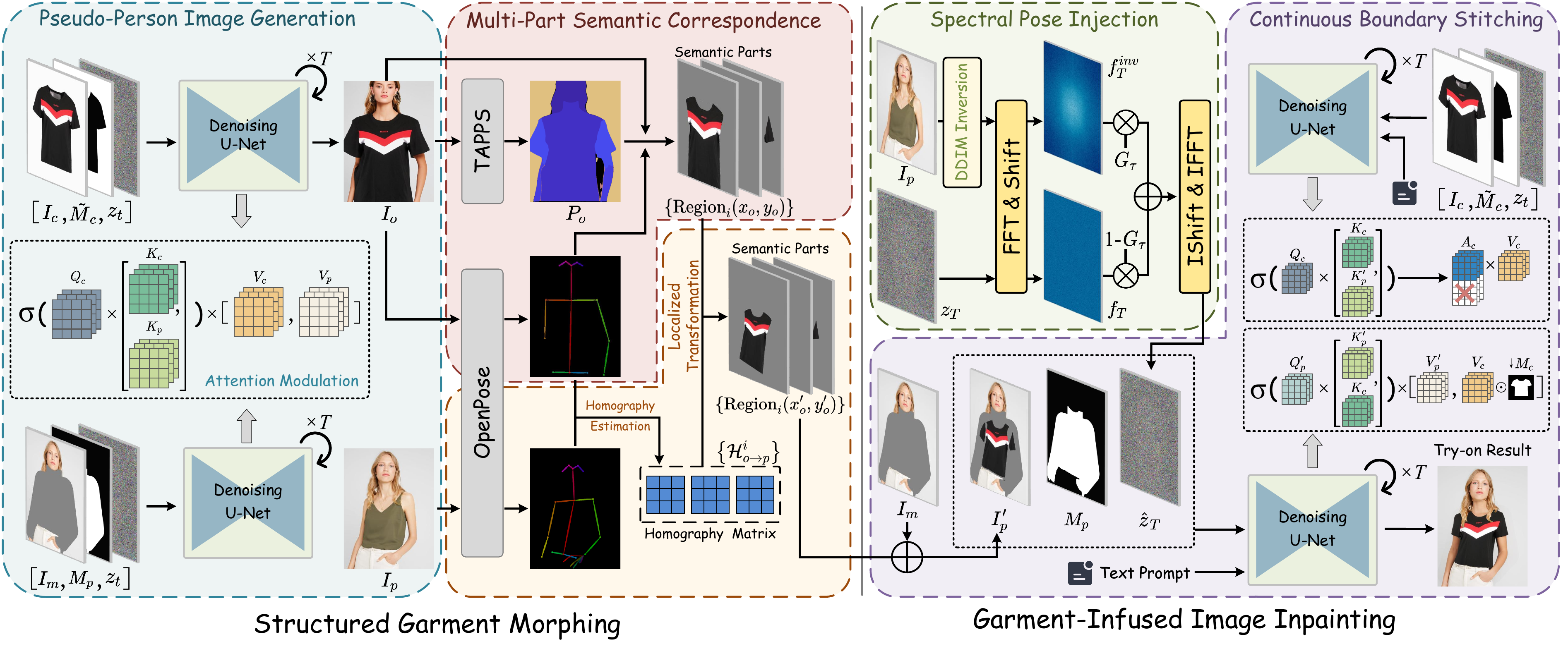}
	\vspace{-8mm}
	\captionof{figure}{Overview of OmniVTON. It consists of two main steps: 1) Utilize pseudo-person image $I_o$ to achieve multi-part deformation, generating adapted clothing. 2) Integrate this prior with clothing-agnostic image $I_m$ to create garment-infused image $I_p^\prime$, which is concatenated with pose-encoded noise $\hat{z}_T$ as input, thereby obtaining refined try-on result through the Continuous Boundary Stitching mechanism.}
	\label{fig:pipeline}
	\vspace{-4mm}
\end{figure*}
\textbf{Garment Warping.}
Garment warping plays a fundamental role in virtual try-on by ensuring precise human-body alignment and texture preservation. Early approaches~\cite{viton,vitonhd} relied on Thin Plate Spline~\cite{tps} (TPS) deformation with sparse control points, but their low-dimensional parametric representation struggled to accommodate complex pose variations. Later works~\cite{clothflow,styleflow,gp-vton,d4-vton} improved deformation quality by predicting dense optical flow~\cite{flow} for pixel-level semantic alignment, though they remained reliant on paired training data. To mitigate data constraints, PastaGAN++~\cite{pastagan++} introduced a patch-routed disentanglement module for unpaired training. However, existing methods are often tailored to specific scenarios, limiting their adaptability across diverse input sources.
In contrast, our OmniVTON enables training-free, universal garment adaptation by leveraging skeletal guidance. While StreetTryOn~\cite{street-tryon} also supports cross-scenario applications, its dense warping mechanism struggles with lower garments in in-shop scenarios and fails to preserve garment integrity. Our approach overcomes these limitations, achieving superior versatility and fidelity across a wide range of virtual try-on tasks.

\noindent\textbf{Image-Based Virtual Try-On.}
Implicit warping-based virtual try-on methods~\cite{cat-dm,stableviton,idm-vton,ladi-vton,ootd,wear-any-way} have recently gained attention for their ability to jointly model garment deformation and human-body synthesis using diffusion models’ powerful semantic correspondence capabilities. Ladi-VTON~\cite{ladi-vton}, for instance, employs textual inversion~\cite{textual-inversion} to map garment textures to text-based conditions, but its reliance on textual ambiguity results in insufficient control over garment details. To improve garment-body interactions, IDM-VTON~\cite{idm-vton} and StableVITON~\cite{stableviton} incorporate advanced attention mechanisms, yet the absence of explicit deformation constraints often leads to geometric misalignment and texture inconsistencies, particularly in open-domain scenarios. 
Unlike these methods, OmniVTON provides direct texture guidance through structured garment priors during the inpainting stage, ensuring precise alignment and preserving fine-grained garment details. Its training-free paradigm enables universal applicability across diverse datasets, scenarios, and garment categories.

\noindent\textbf{Exemplar-Based Image Inpainting.}
Both exemplar-based image inpainting~\cite{anydoor,tigic,pbe} and virtual try-on require accurate feature transfer from reference images to target regions. PBE~\cite{pbe} trains an image encoder to align visual and textual semantics, while AnyDoor~\cite{anydoor} enhances texture representation by injecting multi-level high-frequency features into U-Net~\cite{u-net}. However, excessive high-frequency retention often causes style inconsistencies in the generated outputs. In contrast, text-driven inpainting methods~\cite{brushnet,powerpaint} suffer from limited information granularity, leading to texture distortion, whereas personalized approaches~\cite{textual-inversion,e4t} generate more identity-preserving text embeddings but require fine-tuning. 
Our proposed Spectral Pose Injection (SPI) offers a novel alternative by abandoning strict high-frequency constraints while leveraging OmniVTON to maintain garment identity consistency. By integrating structured pose-aware noise modulation, SPI ensures that try-on results conform precisely to the target person's pose without sacrificing texture fidelity.

\section{Approach}
\label{sec:method}
Given a garment-contained\footnote{A garment-contained image $I_c$ refers to either a standalone garment or a person wearing it in diverse backgrounds.} image $I_c$ and a target person image $I_p$, our goal is to seamlessly transfer the indicated garment onto the corresponding semantic region of $I_p$ without any training. 
To this end, we tackle two key challenges:
\begin{itemize}
    \item Warping the given garment in a training-free manner.
    \item Preserving the original person's pose while inpainting the cloth-agnostic image, also without training.
\end{itemize}

As illustrated in Fig.~\ref{fig:pipeline}, OmniVTON follows a two-step workflow. First, it morphs the target garment to create a garment prior aligned with the human body. Then, using this prior and pose-encoded noise, it progressively refines the boundary of the garment and completes the garment-infused image, ensuring a coherent and pose-matching result. 
\subsection{Structured Garment Morphing}
\label{sec:sgm}
We propose Structured Garment Morphing (SGM) to accurately deform the target garment. Unlike TPS and Flow-based methods~\cite{vitonhd,gp-vton}, which require retraining for different domains, SGM leverages skeletal information and parsing maps to constrain garment morphing. It establishes a one-to-one mapping between the target garment and the original worn person images using correspondences between the target and source person. While this approach is naturally applicable in Non-Shop-to-X settings, the Shop-to-X setting faces challenges when only the garment image is available, leading to failures in keypoint detection and parsing due to the lack of parseable human body structure. To ensure universality, SGM's first task is to generate a pseudo-person image from the garment to extract reliable information.

\noindent\textbf{Pesudo-Person Image Generation.} We empirically found that text-driven image generation paradigm often fails to produce the desired pseudo-human image $I_o$, likely due to weak controllability or the need for carefully crafted prompts. Instead, we propose modulating attention outputs for generation. Specifically, we first relocate the target garment to the semantically relevant region of the source person’s body, establishing an initial spatial correspondence. To inject the person's semantic features, we parallel-denoise both the garment-conditioned noise $z_t$, concatenated with the garment image $I_c$ and the inverted cloth mask $\tilde{M}_c$, which indicates the region to be generated. We also apply the same noise conditioned on the worn person's information, concatenated with the cloth-agnostic image $I_m$ and the agnostic mask $M_p$, integrating the latter’s key and value into the self-attention layers of the former:
\begin{equation}
f_{c} = \operatorname{Softmax}\left(\frac{Q_{c} \cdot [K_{c} \parallel K_{p}]^\top}{\sqrt{d}}\right)[V_{c} \parallel V_{p}],
\end{equation}
where $K_p$ and $V_p$ are the key and value matrices of the worn person image, $Q_c$, $K_c$, and $V_c$ represent the query, key, and value of the garment image, and $\parallel$ denotes tensor concatenation along the spatial dimension. Since the key and value encode an image’s spatial layout and content information~\cite{tewel2023key}, this mechanism effectively incorporates human body semantics into the pseudo-person generation process while preserving the target garment’s texture.

\noindent\textbf{Multi-Part Semantic Correspondence.}
After obtaining the target person image, we use skeleton and parsing to establish multi-part semantic correspondence between the target garment and the original worn person image. First, we define a set of $N$ human semantic regions and use OpenPose~\cite{openpose} for semantic disentanglement on both images. 

Taking an upper garment as an example, we define five semantic regions: torso, left and right upper arms, and left and right lower arms. Using the 25 keypoints predicted by OpenPose, we construct bounding boxes $\{B^i_o\}_{i=1}^5$ to encompass all keypoints of each region in $I_o$. Likewise, we obtain the corresponding bounding boxes $\{B_p^i\}_{i=1}^5$ for $I_p$.

To avoid interference from overlapping parts, we then apply a human part segmentation map $P_o$, generated by TAPPS~\cite{tapps}, to isolate pixels corresponding to each region:
\begin{equation}
\mathbb{I}_{\text{Region}_i}(x_o,y_o)=\left\{
\begin{array}{ll}
    1,& \text{if }(x_o,y_o)\in P_o^i\cap B_o^i,\\
    0,& \text{otherwise.}
\end{array}
\right.
\end{equation}
where $\mathbb{I}(\cdot)$ indicates the indicator function, and $(x_o,y_o)$ represents the pixel coordinates of $I_o$. Note that although the generated pseudo-person image may not always capture the entire human body, our relocate operation ensures that at least the outer body, including the garment, is covered. This is sufficient to establish the multi-part correspondence needed for the subsequent localized transformation.

\noindent\textbf{Localized Transformations.} For the corner points of each bounding box pair $\{B^i_o,B^i_p\}$, we optimize the homography matrix $\mathcal{H}^i_{o\rightarrow p}\in\mathbb{R}^{3\times 3}$ using the Levenberg-Marquardt algorithm~\cite{levenberg}. Then, a piecewise perspective transformation is applied to $I_o$ to align it with the source human geometry:
\begin{equation}
\begin{bmatrix}
x_o'\\
y_o'\\
1
\end{bmatrix}
=\sum_{i=1}^{5} \mathbb{I}_{\text{Region}_i(x_o,y_o)}H_{o\rightarrow p}^i
\begin{bmatrix}
x_o\\
y_o\\1
\end{bmatrix},
\end{equation}
where $(x_o',y_o')$  represents the pixel coordinates at $(x_o,y_o)$ after morphing. In this way, we obtain a coarse deformed garment $I_\omega$, which, through multi-region stitching, serves as an effective prior for the subsequent step of garment-infused image inpainting.

The one-to-one mapping characteristic of SGM eliminates the need for training; however, this multi-region stitching approach results in discontinuities along the boundaries of the morphed garment. We will discuss methods for boundary refinement in Sec.~\ref{sec:cbs}.

\subsection{Spectral Pose Injection}
\label{sec:SPI}
The VTON task requires high human pose fidelity, as relying solely on the local deformation from garment morphing is insufficient for full-body pose alignment. This issue is especially noticeable with garments of ambiguous structures, like sleeveless vests or shorts. While skeleton-based conditioning in diffusion models can improve pose controllability, combining it with other conditions, such as text prompts, may cause the model to overfit certain conditions while neglecting others~\cite{emma}. Alternatively, we apply DDIM Inversion~\cite{ddim} to reverse-map $I_p$ into latent space, obtaining noise $z_T^{inv}$ that preserves source human body structure. However, $z_T^{inv}$ also retains the source garment’s texture, which may conflict with the texture generation of the target garment during image inpainting.

\begin{figure}[t]
	\centering
	\includegraphics[width=.9\linewidth]{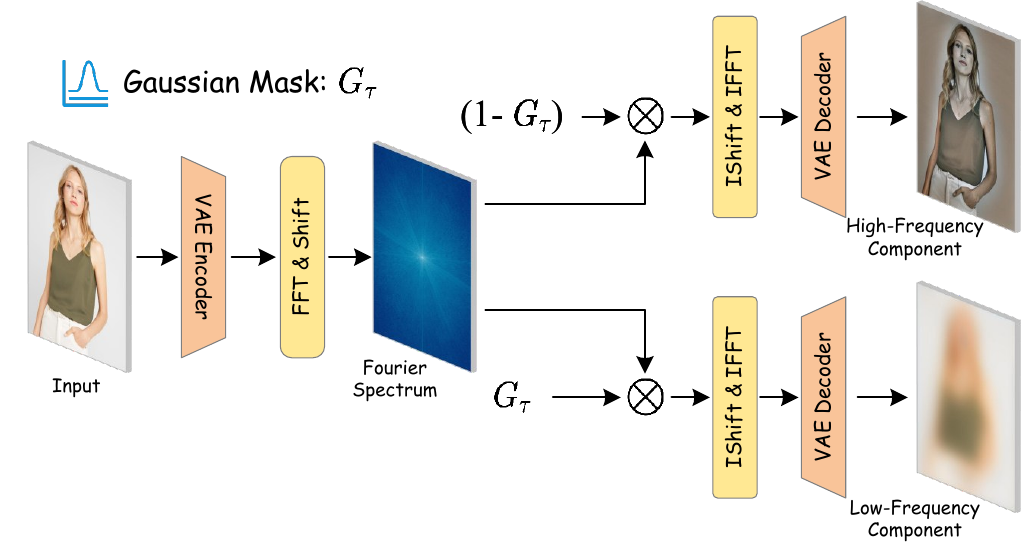}
	\vspace{-3mm}
	\captionof{figure}{Visualization of distinct spectral bands in latent space.}
	\label{fig:frequency}
	\vspace{-4mm}
\end{figure}

To address this, we propose Spectral Pose Injection (SPI), inspired by our spectral analysis in latent space. As shown in Fig.~\ref{fig:frequency}, the human latent, when decomposed via the Fast Fourier Transform (FFT) into low- and high-frequency components, exhibits distinct characteristics in reconstructing the original image. The low-frequency component captures only the coarse human silhouette, which adequately preserves pose information, while the high-frequency component retains both pose and fine garment textures.

The core idea of SPI is to retain the low-frequency structural information from the inverted noise while leveraging the high-frequency components of random noise to enhance generative flexibility. Specifically, we first apply FFT and centralization to both $z_T^{inv}$ and a random noise $z_T$:
\begin{equation}
f_T^{inv} = \text{Shift}(\mathcal{F}(z_T^{inv})), \quad f_T = \text{Shift}(\mathcal{F}(z_T)),
\end{equation}
where $\mathcal{F}(\cdot)$ denotes the Fourier transform, and $\text{Shift}(\cdot)$ shifts the low-frequency components to the center, facilitating spectral decoupling.

Next, we perform frequency-domain weighted fusion using a Gaussian low-pass mask $G_\tau$, where $\tau$ controls the cutoff frequency:
\begin{equation}
\hat{f}_T = G_\tau \odot f_T^{inv} + (1 - G_\tau) \odot f_T,
\end{equation}
where $\odot$ denotes element-wise multiplication. The mask $G_\tau$  ensures that the low-frequency pose information from $z_T^{inv}$ is preserved while injecting the high-frequency randomness from $z_T$ to eliminate texture residuals. 

Finally, we apply the inverse Fourier transform to the fused spectrum $\hat{f}_T$ to obtain the mixed initial noise:
\begin{equation}
\hat{z}_T = \mathcal{F}^{-1}(\text{Shift}^{-1}(\hat{f}_T)).
\end{equation}
\subsection{Continuous Boundary Stitching}
\label{sec:cbs}
During the inpainting stage, the diffusion model takes the concatenation of mixed noise $\hat{z}_T$, the garment-infused image $I_p^\prime$, and the cloth-agnostic mask as input to generate the final try-on image. Since $I_\omega$ is assembled by morphing and stitching garment regions, its boundaries may exhibit texture discontinuities. These artifacts can be misinterpreted by the diffusion model as inherent garment details, resulting in unrealistic seams or misaligned patterns in the output.

To address this, we propose the Continuous Boundary Stitching (CBS) mechanism, which leverages bidirectional semantic context information between $I_c$ and $I_p^\prime$ to improve boundary continuity during the inpainting process. Similar to attention modulation in pseudo-person image generation, CBS operates by manipulating the self-attention outputs. The key difference is that CBS enables dual-path feature exchange, where the interaction from the $I_c$-path to the $I_p^\prime$-path is defined as follows:
\begin{equation}\small
f_p^\prime = \operatorname{Softmax}\left(\frac{Q_{p}^\prime \cdot [K_{p}^\prime \parallel K_{c}]^\top}{\sqrt{d}}\right)[V_{p}^\prime \parallel (V_{c}\cdot \downarrow M_c)],\label{eqn:p}
\end{equation}
where $\downarrow$ represents downsampling $M_c$ to match the dimension of $V_c$, aiming to suppress interference from the background information of $I_c$. The operation in Eq.~(\ref{eqn:p}) allows the query $Q_p^\prime$ to match the target garment texture, thereby bridging discontinuities caused by multi-region stitching.

In addition, we also adjust the self-attention output of $I_c$ by using the key from the $I_p^\prime$-path:
\begin{equation}\small
A_c = \operatorname{Softmax}\left(\frac{Q_c\cdot[K_c\parallel K_p^\prime]^\top}{\sqrt{d}}\right), f_c = A_c[:, 1:n]\cdot V_c,
\end{equation}
where $A_c\in\mathbb{R}^{n\times 2n}$ denotes the self-attention map. This operation enhances the similarity between the attention maps of $I_c$ and $I_p^\prime$, while suppressing dissimilar values. As a result, $A_c$ retains its continuous boundary and adjusts to align with the layout of $I_p^\prime$. This optimization further improves the information flow from the $I_c$-path to the $I_p^\prime$-path, aiding boundary refinement in the subsequent time step. Note that we exclude $V_p^\prime$ to prevent texture interference from its discontinuities, so only the first $n$ columns of the attention map are used in the output calculation.

\section{Experiments}
\label{sec:exp}\setcounter{footnote}{0}

\subsection{Experimental Setup}
\noindent\textbf{Datasets.}
We evaluate OmniVTON on two in-shop datasets (VITON-HD~\cite{vitonhd}, DressCode~\cite{dresscode}) and one in-the-wild dataset (DeepFashion2~\cite{deepfashion2}). VITON-HD provides 2,032 upper garment-model test pairs, while DressCode spans three subcategories (upper, lower, and dresses) with a total of 5,400 test samples. For DeepFashion2, following the StreetTryOn benchmark~\cite{street-tryon}, we constructed a 2,089-image test set covering four try-on scenarios: Shop-to-Street, Model-to-Model, Model-to-Street, and Street-to-Street. Input resolution was dynamically adjusted based on the target person source: 512$\times$384 for VITON-HD/DressCode and 512$\times$320 for DeepFashion2.

\noindent\textbf{Baselines and Metrics.}
We compare OmniVTON against two baseline categories: exemplar-based image editing methods (PBE~\cite{pbe}, AnyDoor~\cite{anydoor}, TIGIC~\cite{tigic}, Cross-Image~\cite{cross-image}) and traditional virtual try-on models (PWS~\cite{pws}, PastaGAN++~\cite{pastagan++}, GP-VTON~\cite{gp-vton}, CAT-DM~\cite{cat-dm}, D$^4$-VTON~\cite{d4-vton}, IDM-VTON~\cite{idm-vton}, and StreetTryOn~\cite{street-tryon}). Among image editing methods, PBE and AnyDoor leverage large-scale pretraining for image inpainting, while TIGIC and Cross-Image utilize cross-image attention for localized editing. Traditional VTON models, except StreetTryOn, are scenario-specific: PWS and PastaGAN++ are trained on Model-to-Model datasets (DeepFashion~\cite{deepfashion} and UPT~\cite{pastagan}, respectively), whereas GP-VTON, CAT-DM, and others focus on Shop-to-Model settings.

Evaluation follows standard protocols, using Fréchet Inception Distance (FID)~\cite{fid} to measure the similarity between generated try-on results and real image distributions. For VITON-HD and DressCode, which contain ground-truth images, we also employ Structural Similarity (SSIM)~\cite{ssim} and Learned Perceptual Image Patch Similarity (LPIPS)~\cite{lpips} to assess structural integrity and texture fidelity.

\subsection{Comparison with State-of-the-Art Methods}
\noindent\textbf{Quantitative Evaluation.}
Tab.~\ref{tab:vitonhd} presents the quantitative evaluation of OmniVTON on the VITON-HD dataset. To assess \textit{\textbf{cross-dataset}} generalization, all VTON methods were tested using official checkpoints pre-trained on DressCode. OmniVTON outperforms the best-performing baseline by 0.020 in SSIM and 0.002 in LPIPS, confirming its superiority in pose preservation and appearance fidelity. More notably, our method reduces the FID metric by 5.209 in the unpaired setting, demonstrating exceptional cross-domain adaptability. Notably, while the second-best performer, AnyDoor, benefits from training on a dataset that includes VITON-HD samples, leading to favorable FID$_u$ and FID$_p$ scores, its structural accuracy remains constrained by the absence of geometric garment guidance.

\begin{table}[t]\centering
\def\arraystretch{1.2}
\small
\scriptsize
\tabcolsep 1.5pt
\resizebox{\linewidth}{!}{
    \begin{tabular}{lc cc ccccc}
        \toprule
        \multicolumn{2}{c}{Method} && Year && FID$_u$$\downarrow$ & FID$_p$$\downarrow$ & SSIM$_p$$\uparrow$ & LPIPS$_p$$\downarrow$ \\
        \cmidrule{1-2} \cmidrule{4-4} \cmidrule{6-9}
        \multicolumn{2}{c}{PBE~\cite{pbe}} && 2023 (CVPR) && 19.230 & 17.649 & 0.784 & 0.227 \\
        \multicolumn{2}{c}{AnyDoor~\cite{anydoor}} && 2024 (CVPR) && 14.830 & 9.922 & 0.796 & 0.164 \\
        \multicolumn{2}{c}{TIGIC~\cite{tigic}} && 2024 (ECCV) && 90.338 & 88.900 & 0.613 & 0.422 \\
        \multicolumn{2}{c}{Cross-Image~\cite{cross-image}} && 2024 (SIGGRAPH) && 62.614 & 57.286 & 0.760 & 0.256 \\
        \cmidrule{1-2} \cmidrule{4-4} \cmidrule{6-9}
        \multicolumn{2}{c}{GP-VTON~\cite{gp-vton}} && 2023 (CVPR) && 51.566 & 49.196 & 0.810 & 0.249 \\
        \multicolumn{2}{c}{CAT-DM~\cite{cat-dm}} && 2024 (CVPR) && 28.869 & 26.339 & 0.775 & 0.229 \\
        \multicolumn{2}{c}{D$^4$-VTON~\cite{d4-vton}} && 2024 (ECCV) && 25.299 & 23.914 & 0.790 & 0.250 \\
        \multicolumn{2}{c}{IDM-VTON~\cite{idm-vton}} && 2024 (ECCV) && 23.035 & 20.460 & 0.812 & 0.147 \\
        \cmidrule{1-2} \cmidrule{4-4} \cmidrule{6-9}
        \multicolumn{2}{c}{Ours} && - && \textbf{9.621} & \textbf{7.758} & \textbf{0.832} & \textbf{0.145}\\ 
        \bottomrule
    \end{tabular}
}
\vspace{-2mm}
\captionof{table}{Quantitative comparisons on the VITON-HD dataset~\cite{vitonhd}, where the subscript $u$ and $p$ indicates the unpaired and paired settings, respectively.}
\vspace{-2mm}
\label{tab:vitonhd}
\end{table}

\begin{table}[t]\centering
\def\arraystretch{1.2}
\small
\scriptsize
\tabcolsep 1.5pt
\resizebox{\linewidth}{!}{
    \begin{tabular}{lc cc ccccc}
        \toprule
       \multicolumn{2}{c}{Method} && Year && FID$_u$$\downarrow$ & FID$_p$$\downarrow$ & SSIM$_p$$\uparrow$ & LPIPS$_p$$\downarrow$ \\
        \midrule
        \cmidrule{1-2} \cmidrule{4-4} \cmidrule{6-9}
        \multicolumn{2}{c}{PBE~\cite{pbe}} && 2023 (CVPR) && 14.851 & 13.677 & 0.846 & 0.155 \\
        \multicolumn{2}{c}{AnyDoor~\cite{anydoor}} && 2024 (CVPR) && 14.562 & 14.411 & 0.798 & 0.202 \\
        \multicolumn{2}{c}{TIGIC~\cite{tigic}} && 2024 (ECCV) && 64.117 & 63.531 & 0.749 & 0.319 \\
        \multicolumn{2}{c}{Cross-Image~\cite{cross-image}} && 2024 (SIGGRAPH) && 38.438 & 34.917 & 0.841 & 0.161 \\
        \cmidrule{1-2} \cmidrule{4-4} \cmidrule{6-9}
        \multicolumn{2}{c}{GP-VTON~\cite{gp-vton}} && 2023 (CVPR) && 44.753 & 44.469 & 0.843 & 0.218 \\
        \multicolumn{2}{c}{CAT-DM~\cite{cat-dm}} && 2024 (CVPR) && 13.678 & 12.028 & 0.858 & 0.125 \\
        \multicolumn{2}{c}{D$^4$-VTON~\cite{d4-vton}} && 2024 (ECCV) && 22.390 & 21.435 & 0.841 & 0.152 \\
        \multicolumn{2}{c}{IDM-VTON~\cite{idm-vton}} && 2024 (ECCV) && 9.685 & 8.377 & 0.842 & 0.138 \\
        \cmidrule{1-2} \cmidrule{4-4} \cmidrule{6-9}
        \multicolumn{2}{c}{Ours} && - && \textbf{6.450} & \textbf{5.335} & \textbf{0.865} & \textbf{0.119} \\
        \bottomrule
    \end{tabular}
}
\vspace{-2mm}
\captionof{table}{Quantitative comparisons on the DressCode dataset~\cite{dresscode}, where the subscript $u$ and $p$ indicates the unpaired and paired settings, respectively.}
\vspace{-2mm}
\label{tab:dresscode}
\end{table}

\begin{table}[t]
\centering
\def\arraystretch{1.2}
\small
\scriptsize
\tabcolsep 1.5pt
\resizebox{\linewidth}{!}{
\begin{tabular}{lc cc cc cc c}
\toprule
 && Shop-to-Street && Model-to-Model && Model-to-Street && Street-to-Street \\
 \cmidrule{3-3} \cmidrule{5-5} \cmidrule{7-7} \cmidrule{9-9}
&& FID$\downarrow$ && FID$\downarrow$ && FID$\downarrow$ && FID$\downarrow$ \\
\cmidrule{1-1} \cmidrule{3-3} \cmidrule{5-5} \cmidrule{7-7} \cmidrule{9-9}
     
PBE~~\cite{pbe} && 81.538 && 20.181 && 62.664 && 36.556 \\ 
AnyDoor~~\cite{anydoor} && 50.893 && 24.235 && 51.861 && 35.139 \\ 
TIGIC~~\cite{tigic} && 100.177&& 114.151 && 130.836 && 121.520 \\ 
Cross-Image~~\cite{cross-image} && 69.444 && 52.310 && 66.755 && 57.753 \\ 
\cmidrule{1-1} \cmidrule{3-3} \cmidrule{5-5} \cmidrule{7-7} \cmidrule{9-9}
CAT-DM~~\cite{cat-dm} && 37.484 && - && - && - \\ 
D$^4$-VTON~~\cite{d4-vton} && 35.003 && - && - && - \\ 
IDM-VTON~~\cite{idm-vton} && 42.282 && - && - && - \\ 
PWS~~\cite{pws} && - && 34.858 && 77.274 && 84.990 \\
PastaGAN++~~\cite{pastagan++} && - && 13.848 && 71.090 && 67.016 \\
StreetTryOn~~\cite{street-tryon} && 34.054 && 12.185 && 34.191 && 33.039 \\
\cmidrule{1-1} \cmidrule{3-3} \cmidrule{5-5} \cmidrule{7-7} \cmidrule{9-9}
Ours && \textbf{33.919} && \textbf{8.983} && \textbf{33.450} && \textbf{23.470} \\ 
\bottomrule
\end{tabular}
}
\vspace{-2mm}
\captionof{table}{Quantitative comparisons on the StreetTryOn benchmark~\cite{street-tryon}. Virtual try-on methods use publicly available models trained on VITON-HD~\cite{vitonhd}, while PWS~\cite{pws} and PastaGAN++~\cite{pastagan++} are trained on DeepFashion~\cite{deepfashion} and UPT~\cite{pastagan}, respectively. StreetTryOn results are taken from its original paper.
}
\vspace{-4mm}
\label{tab:street_tryon}
\end{table}

\begin{figure*}[t]
	\centering
	\includegraphics[width=\linewidth]{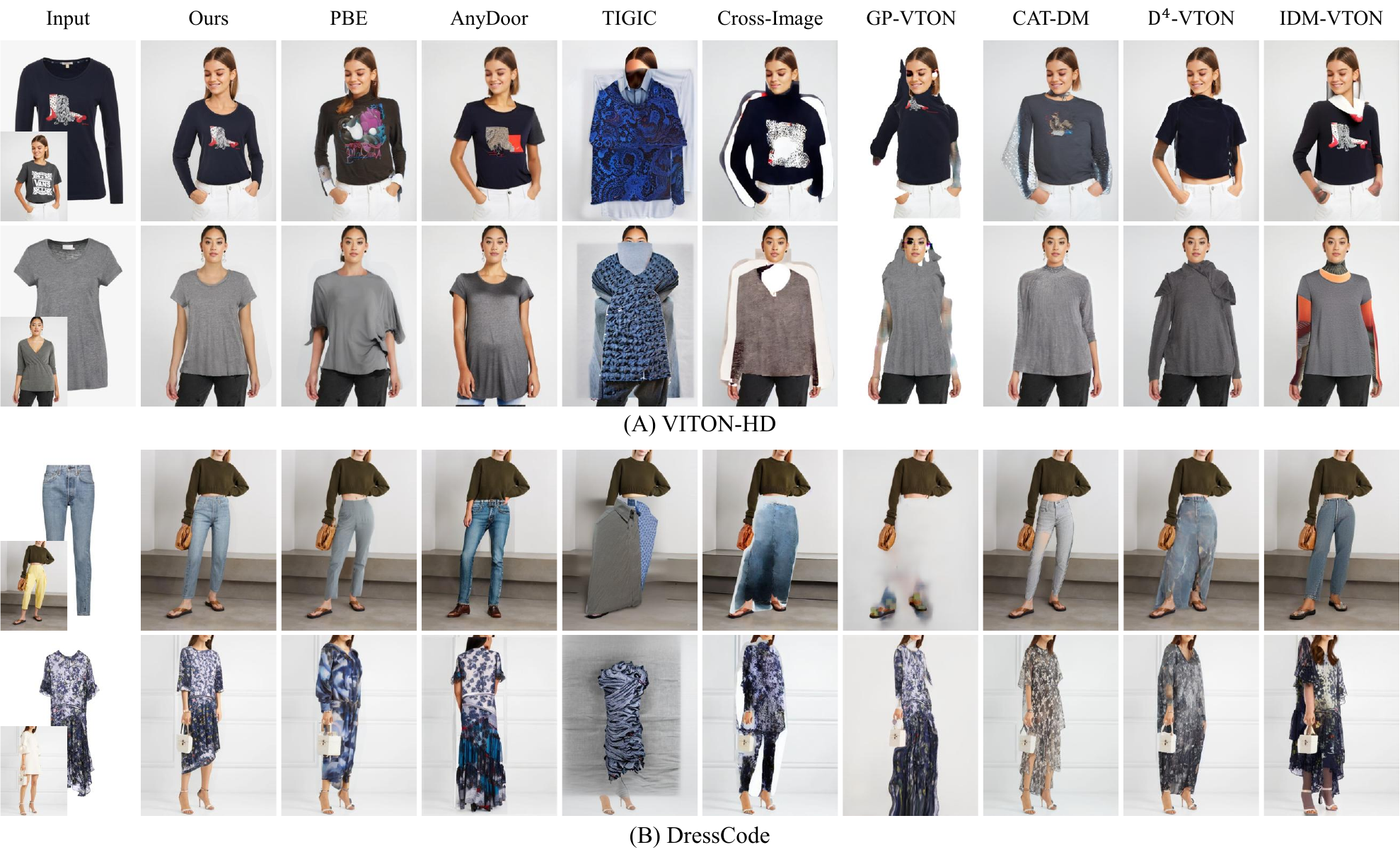}
	\vspace{-6mm}
	\captionof{figure}{Qualitative results across multiple datasets and clothing types. We provide upper garment try-on results on the VITON-HD dataset~\cite{vitonhd} (top) and lower garment/dresses visual comparisons on the DressCode dataset~\cite{dresscode} (bottom).}
	\label{fig:vitonhd_dresscode}
	\vspace{-5mm}
\end{figure*}

\begin{figure}
	\centering
	\includegraphics[width=\linewidth]{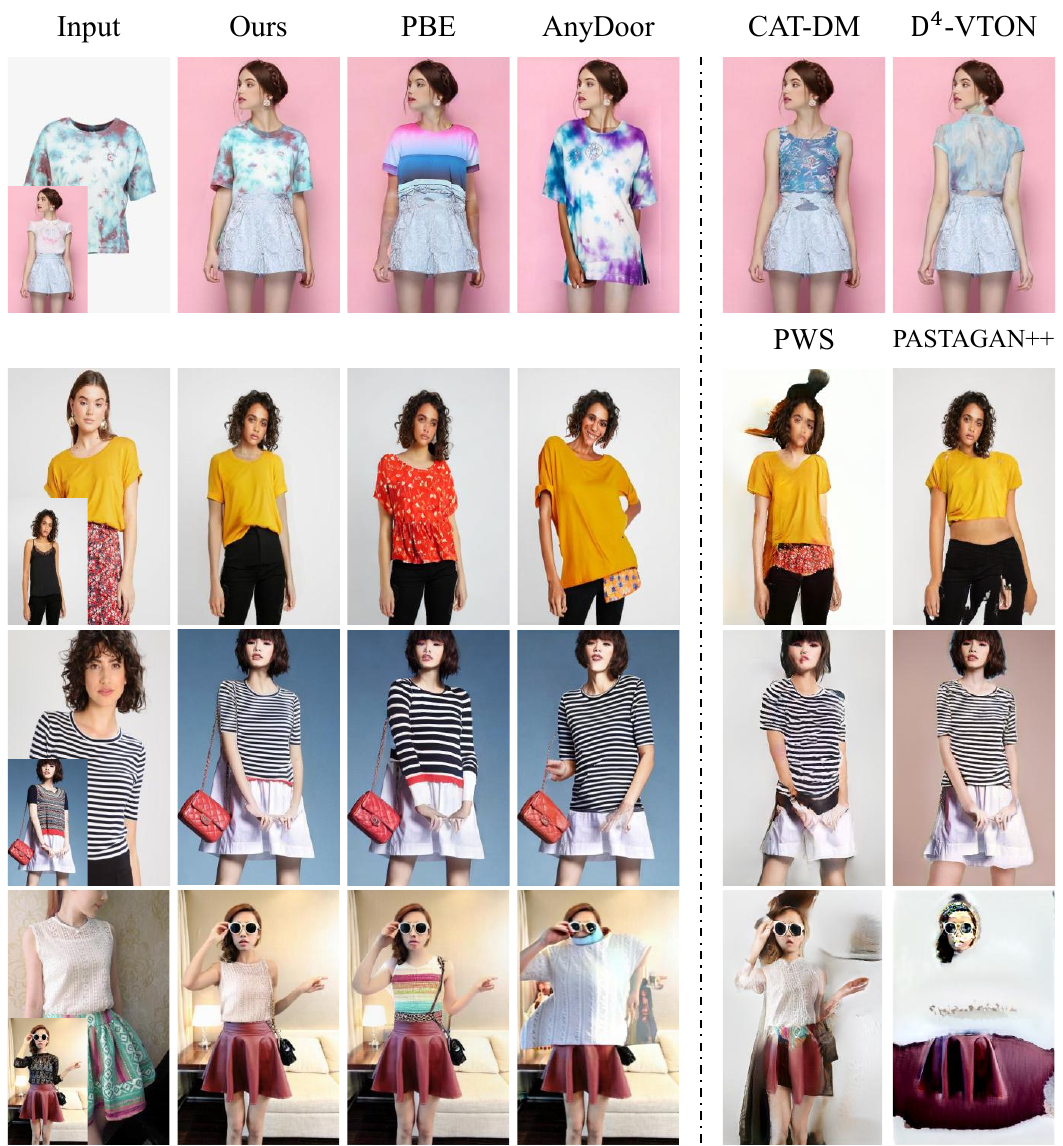}
	\vspace{-6mm}
	\captionof{figure}{Qualitative results on the StreetTryOn benchmark~\cite{street-tryon} under different scenarios (from top to bottom): Shop-to-Street, Model-to-Model, Model-to-Street, and Street-to-Street.}
	\label{fig:street}
	\vspace{-4mm}
\end{figure}

To evaluate \textit{\textbf{cross-type}} adaptability, we tested VTON methods using VITON-HD pre-trained models (which contain only upper garments) on the DressCode dataset, which includes diverse clothing types. As shown in Tab.~\ref{tab:dresscode}, OmniVTON achieves substantial improvements across all metrics, outperforming both exemplar-based editing and VTON baselines with at least a 33.4\% relative enhancement in FID$_u$. This performance gain stems from the synergistic effects of Structured Garment Morphing (SGM) and Continuous Boundary Stitching (CBS), which collectively enhance robustness across varied garment styles.

Beyond Shop-to-Model scenarios, Tab.~\ref{tab:street_tryon} systematically compares \textit{\textbf{cross-scenario}} try-on performance. Missing entries (`-') denote scenario-specific limitations of certain methods\footnote{We exclude GP-VTON due to unavailable parsing models.}. In Shop-to-Street tasks, D$^4$-VTON and StreetTryOn, benefiting from warping priors, outperform prior-free methods, yet OmniVTON surpasses them in body reconstruction through Spectral Pose Injection (SPI). In Model-to-Model, Model-to-Street, and Street-to-Street settings, our training-free framework maintains a significant advantage, even outperforming StreetTryOn despite it being trained on in-domain data. Moreover, StreetTryOn struggles with lower-body garments and dresses in Shop-to-Street and Shop-to-Model tasks, as garment DensePose~\cite{garment-densepose} fails to provide reliable predictions for these clothing categories. In contrast, SGM successfully generates pseudo-person images, ensuring comprehensive cross-scenario applicability.

\noindent\textbf{Qualitative Evaluation.}  
We present qualitative results on the VITON-HD and DressCode datasets in Fig.~\ref{fig:vitonhd_dresscode}. TIGIC and Cross-Image fail to generate realistic human images due to their lack of task-specific designs. While inpainting models (PBE, AnyDoor) and traditional VTON methods improve human-body generation, they fail to transfer garment textures effectively and introduce noticeable artifacts, especially in cross-domain scenarios. As a GAN-based~\cite{gan} method, GP-VTON exhibits poor human-body completion due to its limited generalization capability. In contrast, OmniVTON achieves high-fidelity try-on results while preserving garment textures with precision.

Since StreetTryOn has not released its code, we compare six alternative methods on this benchmark. Among them, CAT-DM and D$^4$-VTON are restricted to in-shop garment inputs, while PWS and PastaGAN++ extract target garments from model images. As shown in Fig.~\ref{fig:street}, although generic inpainting models (PBE, AnyDoor) demonstrate adaptability to different scenarios, they fail to maintain pose and texture consistency between pre- and post-try-on images. Scenario-specific methods like PWS and PastaGAN++, trained on constrained backgrounds, struggle with real-world complexities. OmniVTON consistently achieves accurate garment texture transfer and pose alignment across diverse settings, demonstrating superior generalizability.

\subsection{Ablation Analysis}
We conduct an ablation analysis to assess the contributions of OmniVTON's core modules through four model variants. Starting from a baseline model (Base) using only text prompts, we incrementally integrate: (A) Structured Garment Morphing (SGM) for garment priors, (B) Continuous Boundary Stitching (CBS) for boundary refinement, and (C) Spectral Pose Injection (SPI) for pose-aware noise control. OmniVTON combines all components.

\begin{figure}
	\centering
	\includegraphics[width=\linewidth]{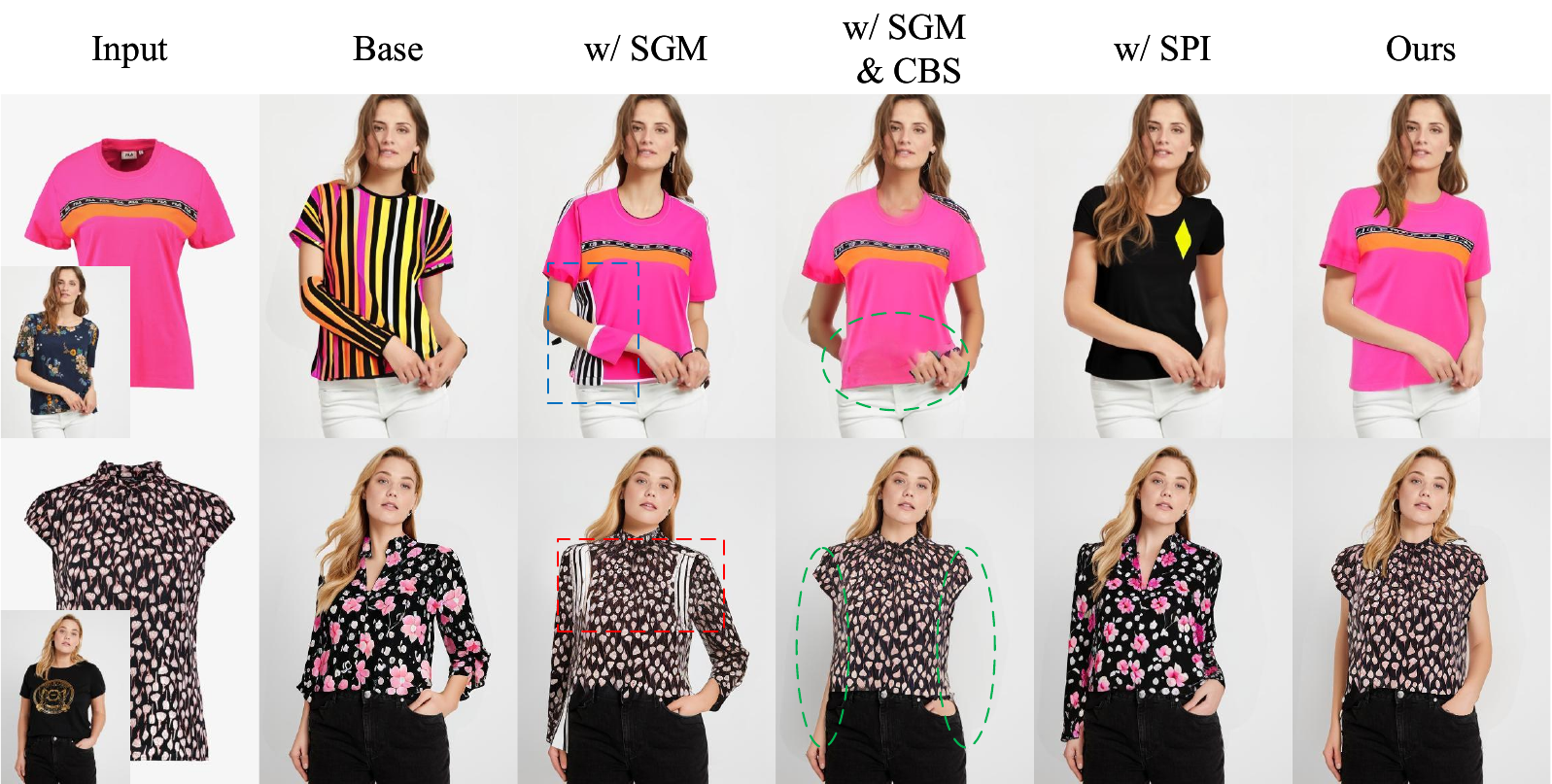}
	\vspace{-6mm}
	\captionof{figure}{Qualitative ablation study on different variants.}
	\label{fig:ablation}
	\vspace{-6mm}
\end{figure}

\noindent\textbf{Effectiveness of SGM.} As shown in Tab.~\ref{tab:ablation}, (A) significantly outperforms Base across all metrics, confirming that SGM effectively aligns garments without end-to-end training. Fig.~\ref{fig:ablation} further illustrates its ability to preserve garment textures via fine-grained geometric cues.

\noindent\textbf{Effectiveness of CBS.} CBS eliminates boundary artifacts (red box, Fig.~\ref{fig:ablation}) and refines textures (blue box). Variant (B) improves LPIPS by 0.019 over (A), validating its role in enhancing perceptual quality. Since CBS primarily refines SGM-derived priors, we focus on their combined effect.

\noindent\textbf{Effectiveness of SPI.} As seen in Tab.~\ref{tab:ablation}, (C) and OmniVTON show notable SSIM and FID$_u$ gains, confirming SPI’s ability to suppress noise contamination while preserving structural consistency. Fig.~\ref{fig:ablation} (green circle) highlights improved body part alignment, reducing pose misalignment. Integrating all components, OmniVTON achieves state-of-the-art texture fidelity and pose consistency.

\begin{table}[t]\centering
\def\arraystretch{1.2}
\small
\scriptsize
\tabcolsep 1.5pt
\resizebox{\linewidth}{!}{
\begin{tabular}{l c c c c c c c c c c}
  \toprule
  \multicolumn{2}{c}{Method} && SGM & CBS & SPI && FID$_u$$\downarrow$ & FID$_p$$\downarrow$ & SSIM$_p$$\uparrow$ & LPIPS$_p$$\downarrow$ \\
  \cmidrule{1-2} \cmidrule{4-6} \cmidrule{8-11} 

  \multicolumn{2}{c}{Base} && & & && 18.445 & 16.878 & 0.773 & 0.222 \\
  \multicolumn{2}{c}{(A)} && \ding{51} & & && 13.303 & 11.475 & 0.809 & 0.177 \\ 
  \multicolumn{2}{c}{(B)} && \ding{51} & \ding{51} & && 9.799 & 7.993 & 0.824 & 0.158 \\ 
  \multicolumn{2}{c}{(C)} && & & \ding{51} && 13.148 & 10.767 & 0.813 & 0.180 \\ 
  \cmidrule{1-2} \cmidrule{4-6} \cmidrule{8-11} 
  \multicolumn{2}{c}{OmniVTON} && \ding{51} & \ding{51} & \ding{51} && \textbf{9.621} & \textbf{7.758} & \textbf{0.832} & \textbf{0.145} \\
  \bottomrule
\end{tabular}
}
\vspace{-2mm}
\caption{Ablation study of the Structured Garment Morphing (SGM), Continuous Boundary Stitching (CBS), and Spectual Pose Injection (SPI) on the VITON-HD dataset~\cite{vitonhd}.}
\vspace{-2mm}
\label{tab:ablation}
\end{table}

\subsection{Multi-Human Virtual Try-On}
\begin{figure}
	\centering
	\includegraphics[width=\linewidth]{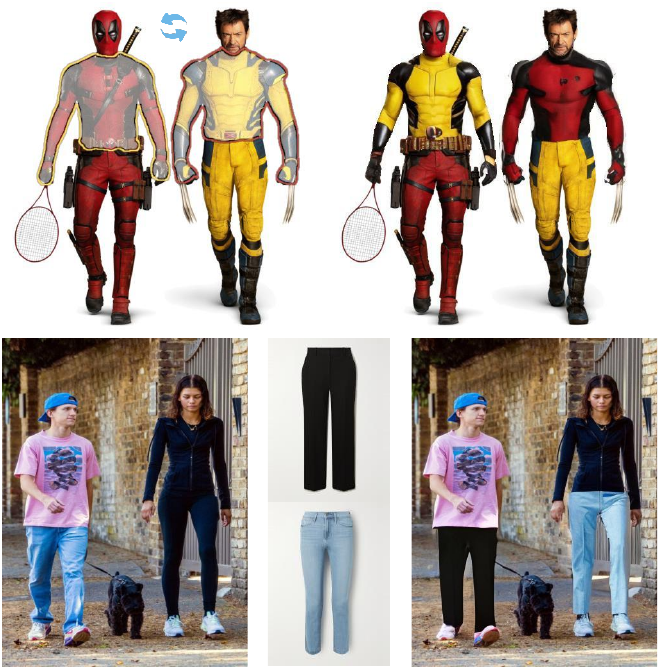}
	\vspace{-6mm}
	\captionof{figure}{Multi-human virtual try-on. Top row shows Model-to-Model, while bottom row depicts Shop-to-Street.}
	\label{fig:multi_user}
	\vspace{-5mm}
\end{figure}

Beyond single-human virtual try-on, our method extends to multi-human interactive group try-on (Fig.~\ref{fig:multi_user}). This capability arises from the innovative design of SGM, which enables seamless adaptation for multiple users. Given multiple garments, we concatenate them along spatial dimensions to generate multiple pseudo-person images simultaneously. By leveraging positional and semantic cues from skeleton and parsing maps, our approach allows for the effortless application of identical or distinct garments to multiple humans. 
Multi-human try-on broadens the scope of virtual try-on tasks, further demonstrating the universality of our method. This extension opens new directions for group-centric fashion experiences, such as coordinated family outfits and uniform design.

\section{Conclusions, Limitations, and Future Work}
In this paper, we present OmniVTON, a training-free universal framework that ensures both texture fidelity and pose consistency across diverse settings. Structured Garment Morphing enables anatomical garment-body alignment, while Continuous Boundary Stitching ensures seamless texture transitions, achieving fine-grained texture consistency without domain-specific training. Spectral Pose Injection further enhances pose alignment through frequency-aware modulation of inversion noise, preserving structural integrity while eliminating texture contamination. Extensive experiments demonstrate OmniVTON’s superiority in flexibility and generalization, particularly in its pioneering capability for multi-human VTON.

Despite its effectiveness, OmniVTON encounters challenges in extreme cases, such as high-density crowds or minimal target body regions, leading to garment misalignment. Visual results illustrating these limitations are provided in the supplementary materials. Future work will focus on developing more robust multi-human try-on frameworks to address these edge cases.

\section*{Acknowledgements}
This work is supported by the National Natural Science Foundation of China (No. 62102381, 41927805); Shandong Natural Science Foundation (No. ZR2021QF035); the National Key R\&D Program of China (No. 2022ZD0117201); the Guangdong Natural Science Funds for Distinguished Young Scholar (No.2023B1515020097); the National Research Foundation, Singapore under its AI Singapore Programme (No.AISG3-GV-2023-011); the Singapore Ministry of Education AcRF Tier 1 Grant (No. MSS25C004); and the Lee Kong Chian Fellowships.
{
    \small
    \bibliographystyle{ieeenat_fullname}
    \bibliography{main}
}

\newpage
\appendix
\maketitlesupplementary

\section{Details and Discussion}
\subsection{Implementation Details}
All experiments were conducted using PyTorch 2.1.1 on a NVIDIA GeForce RTX 3090 GPU. We adopted Stable Diffusion v2~\cite{sd} as the base model, retaining default hyperparameter configurations. For both Pseudo-Person Image Generation and Garment-Infused Image Inpainting, we employed the standard DDIM sampler for deterministic inference with 50 time steps. For Spetral Pose Injection, we set the standard deviation $\tau$ of the Gaussian mask to 0.1.

During the garment morphing stage, we implemented distinct region segmentation strategies for different garment categories: 1) Upper garments underwent five-region processing (left and right upper arms, left and right lower arms, and torso regions); 2) Lower garments were similarly decoupled into five regions (left and right upper legs, left and right lower legs, and hip-above regions); 3) Dresses were segmented into upper and lower garment sections for separate processing. The agnostic and clothing masks are provided by the dataset. In practical applications, SAM~\cite{sam} can be used to obtain the mask corresponding to the user input image.

\subsection{Text Prompts Acquisition}
Here, we describe the process of acquiring text prompts and examine their impact. Specifically, we convert images into text using the CLIP Interrogator~\cite{clipin}, where the generated descriptions consist of a core caption and auxiliary modifier terms. The core caption directly describes the image content, while the auxiliary terms are selected based on cosine similarity between garment features and text embeddings from four predefined datasets: artists, mediums, movements, and flavors.
\begin{figure}
	\centering
	\includegraphics[width=\linewidth]{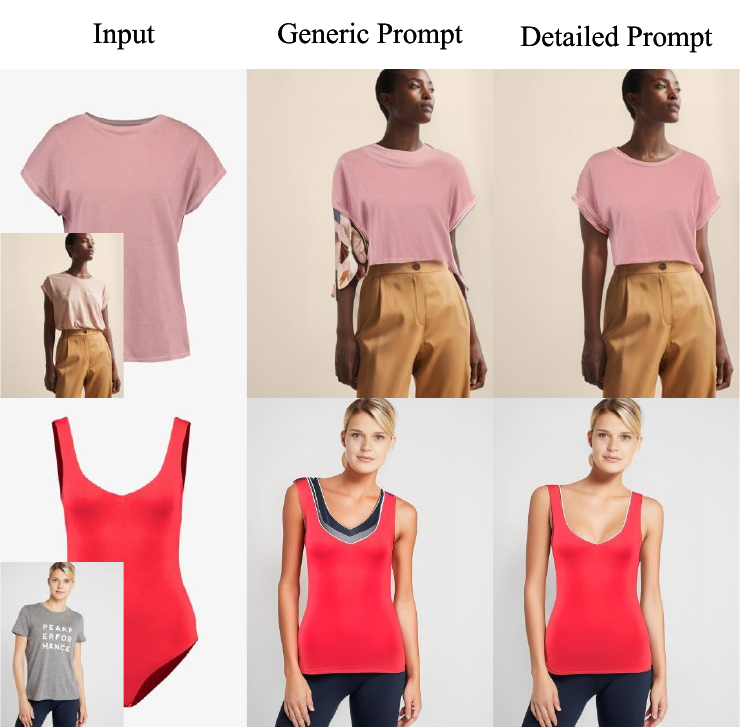}
	\vspace{-6mm}
	\captionof{figure}{Influence of different text prompts.}
	\label{fig:prompt}
	\vspace{-2mm}
\end{figure}

To verify the importance of text prompts in virtual try-on tasks, we conducted a controlled analysis using a generic prompt (``a person wearing an upper garment''). As shown in Fig.~\ref{fig:prompt}, more detailed text prompts lead to try-on results with enhanced identity consistency, highlighting the crucial role of precise textual descriptions in controlling the quality of generation.

\begin{figure}
	\centering
	\includegraphics[width=\linewidth]{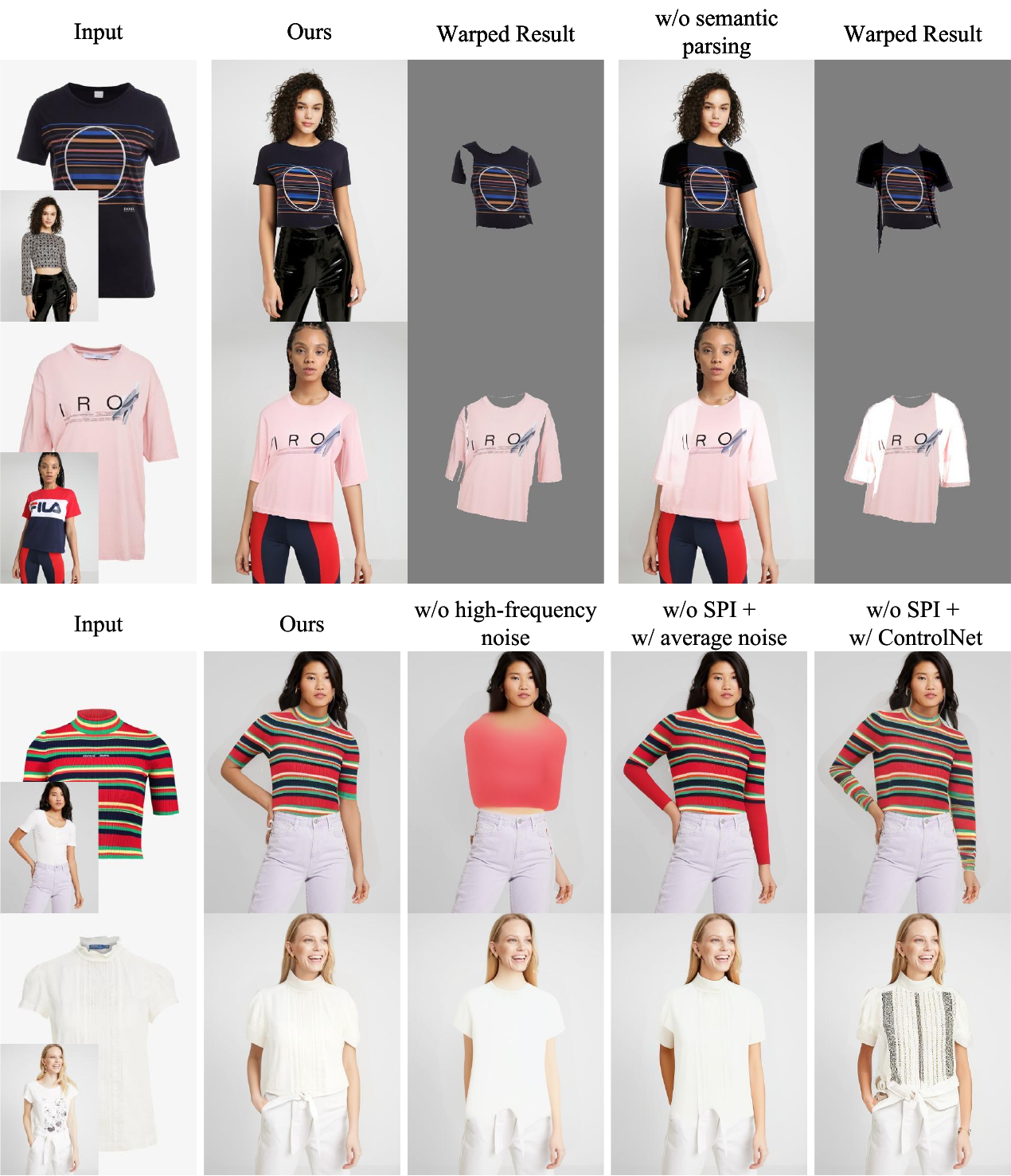}
	\vspace{-6mm}
	\captionof{figure}{Qualitative results of additional ablation analysis.}
	\label{fig:supp_ablation}
	\vspace{-2mm}
\end{figure}

\begin{table}[t]\centering
\def\arraystretch{1.2}
\small
\scriptsize
\tabcolsep 1.5pt
\resizebox{\linewidth}{!}{
\begin{tabular}{l c c c c c c}
  \toprule
  \multicolumn{2}{c}{Method} && FID$_u$$\downarrow$ & FID$_p$$\downarrow$ & SSIM$_p$$\uparrow$ & LPIPS$_p$$\downarrow$ \\
  \cmidrule{1-2} \cmidrule{4-7}
  \multicolumn{2}{c}{OmniVTON} && \textbf{9.621} & \textbf{7.758} & 0.832 & \textbf{0.145} \\
  \multicolumn{2}{c}{w/o semantic parsing} && 13.705 & 11.930 & 0.817 & 0.170 \\ 
  \multicolumn{2}{c}{w/o $I_c$-path attention modulation} && 9.808 & 7.939 & 0.831 & 0.149 \\ 
  \cmidrule{1-2} \cmidrule{4-7}
  
  \multicolumn{2}{c}{w/o high-frequency noise} && 15.817 & 14.558 & 0.836 & 0.182 \\
  \multicolumn{2}{c}{w/o SPI + w/ average noise} && 12.402 & 10.650 & \textbf{0.849} & 0.151 \\ 
  \multicolumn{2}{c}{w/o SPI + w/ ControlNet} && 10.873 & 9.016 & 0.818 & 0.168 \\ 
  \bottomrule
\end{tabular}
}
\vspace{-2mm}
\caption{More ablation studies of different components.}
\vspace{-2mm}
\label{tab:additional_ablation}
\end{table}

\subsection{Additional Ablation Analysis}
To demonstrate the rationale behind our component design, we conducted additional ablation experiments. First, for SGM, the role of semantic parsing is to perform pixel-level segmentation on skeleton-divided semantic regions, enabling multi-part decoupling. As shown in the upper part of Fig.~\ref{fig:supp_ablation}, relying solely on bounding box-based segmentations, without semantic parsing, for localized transformations leads to erroneous morphing and part overlap, significantly degrading the quality of the try-on results. The quantitative comparison of the ``w/o semantic parsing'' setting in Tab.~\ref{tab:additional_ablation} strongly reinforces the necessity of this component. Secondly, the ``w/o $I_c$-path attention modulation'' setting involves replacing the attention modulation in Eq.~(8) of the main paper with the original self-attention mechanism, resulting in noticeable degradation across all evaluation metrics, thus validating the effectiveness of bidirectional semantic context interaction.

For SPI, the lower part of Tab.~\ref{tab:additional_ablation} and Fig.~\ref{fig:supp_ablation} present both quantitative and qualitative results for different variants. The ``w/o high-frequency noise'' variant retains only the low-frequency components of inversion noise, yet the absence of high-frequency noise leads to overly smoothed results. The ``w/o SPI + w/ average noise'' variant averages random noise and inversion noise as the initial noise. Compared with the ``w/o high-frequency noise'' variant, the introduction of random noise significantly improves perceptual quality. However, due to the lack of frequency-domain decoupling, this variant enhances performance in paired settings but fails to suppress source garment texture interference from inversion noise in unpaired settings, causing performance degradation. Furthermore, comparative experiments with ControlNet~\cite{controlnet}-based skeleton-conditioned injection demonstrate that OmniVTON effectively overcomes the inherent bias of diffusion models in handling multiple conditions by decoupling garment and pose constraints, leading to improved try-on results.

In Tab.~\ref{supp_tab:cutoff_ablation}, we provide additional analysis on the sensitivity of the cutoff frequency $\tau$. When $\tau$ is too small, it suppresses low-frequency pose information, limiting SSIM. As $\tau$ increases, metrics generally improve; however, if $\tau$ becomes too large, it preserves excessive high-frequency details, which harms realism and worsens FID. Setting $\tau=0.1$ balances pose consistency and visual fidelity.

\begin{table}[t]\centering
\small
\scriptsize
\resizebox{\linewidth}{!}{
\begin{tabular}{l c c c c c c}
  \toprule
  \multicolumn{2}{c}{$\tau$} && FID$_u$$\downarrow$ & FID$_p$$\downarrow$ & SSIM$_p$$\uparrow$ & LPIPS$_p$$\downarrow$ \\
  \cmidrule{1-2} \cmidrule{4-7}
  \multicolumn{2}{c}{0.01} && 9.941 & 8.185 & 0.823 & 0.160 \\
  \multicolumn{2}{c}{0.05} && 9.620 & 8.033 & 0.829 & 0.153 \\
  \multicolumn{2}{c}{\textbf{0.1}} && 9.621 & 7.758 & 0.832 & 0.145 \\
  \multicolumn{2}{c}{0.3} && 10.056 & 8.150 & 0.842 & 0.140 \\ 
  \multicolumn{2}{c}{0.5} && 11.330 & 9.422 & 0.852 & 0.138 \\ 
  \bottomrule
\end{tabular}
}
\vspace{-2mm}
\caption{Sensitivity analysis of cutoff frequency $\tau$ on VITON-HD.}
\vspace{-2mm}
\label{supp_tab:cutoff_ablation}
\end{table}

\begin{table*}[t]\centering
\small
\scriptsize
\resizebox{\linewidth}{!}{
\begin{tabular}{cccccccccccc}
  \toprule
  &\multicolumn{8}{c}{Training-free} && \multicolumn{2}{c}{Training} \\
  \cmidrule{2-9}\cmidrule{11-12}
  \multirow{3}{*}{Time / Memory} & \multicolumn{2}{c}{OmniVTON} && \multicolumn{2}{c}{TIGIC} && \multicolumn{2}{c}{Cross-Image} && \multicolumn{2}{c}{IDM-VTON} \\
  \cmidrule{2-3}\cmidrule{5-6}\cmidrule{8-9}\cmidrule{11-12}
  & 16.29s & 11,542MB && 13.87s & 23,578MB && 41.49s & 15,748MB && 11.87s & 17,936MB \\
  \midrule
  \specialrule{0em}{1.5pt}{1.5pt}
  \midrule
  \multirow{2}{*}{OmniVTON} & \multicolumn{2}{c}{FID$_u$$\downarrow$} && \multicolumn{2}{c}{SGM Time (s)} && \multicolumn{2}{c}{SPI Time (s)} && \multicolumn{2}{c}{CBS Time (s)} \\
  \cmidrule{2-3}\cmidrule{5-6}\cmidrule{8-9}\cmidrule{11-12}
  & \multicolumn{2}{c}{9.621} && \multicolumn{2}{c}{6.61s} && \multicolumn{2}{c}{3.60s} && \multicolumn{2}{c}{6.08s}\\
  \bottomrule
\end{tabular}
}
\caption{Runtime and memory comparison on VITON-HD.}
\label{supp_tab:cost}
\end{table*}

\subsection{Inference Cost}
As shown in the upper part of Tab.~\ref{supp_tab:cost}, we compare the inference costs of OmniVTON with three state-of-the-art methods. The results show that OmniVTON achieves the lowest memory consumption, outperforms Cross-Image in inference speed, and performs comparably to TIGIC and IDM-VTON, all while maintaining optimal performance. The lower part of the table further presents a module-wise breakdown of inference times. Notably, under the Non-Shop-to-X setting, removing the pseudo-person generation step leads to a sharp reduction in the runtime of the SGM module, from 6.61s to just 0.14s, thereby reducing the overall inference time to 9.82 seconds and further highlighting OmniVTON’s strong potential for real-world deployment.

\subsection{User Study}
\begin{figure}[t]
	\centering
	\includegraphics[width=0.95\linewidth]{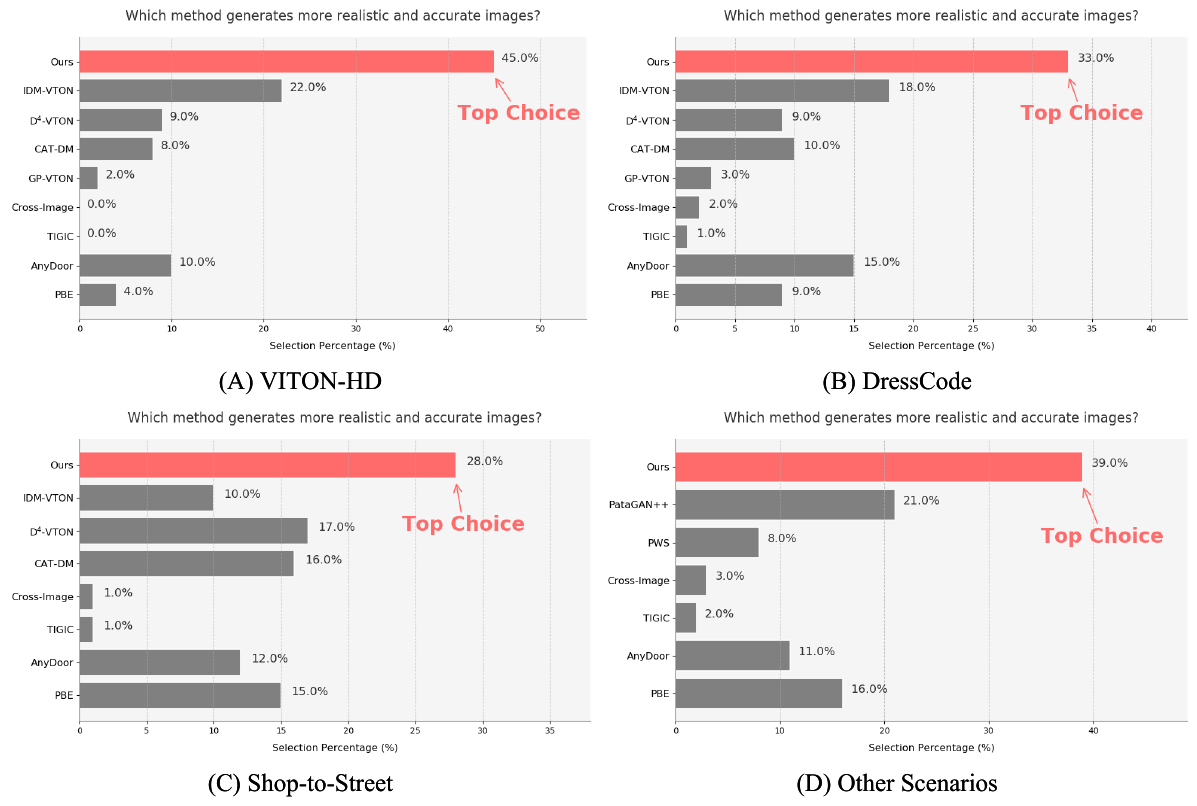}	
	\captionof{figure}{User study on the VITON-HD dataset~\cite{vitonhd}, DressCode dataset~\cite{dresscode} and StreetTryOn benchmark~\cite{street-tryon}.}
	\label{fig:user_study}
	\vspace{-2mm}
\end{figure}

We validate the effectiveness of our method through a rigorously designed user study, establishing a systematic evaluation framework across three benchmark datasets: VITON-HD~\cite{vitonhd}, DressCode~\cite{dresscode}, and StreetTryOn~\cite{street-tryon}. The experiment involved 100 volunteers, each participating in a visual evaluation questionnaire containing 100 comparative sample groups. Specifically, the VITON-HD dataset includes 20 test sample groups, the DressCode dataset covers 40 sample groups across three garment categories (upper, lower, dresses), and the StreetTryOn benchmark allocates the remaining 40 sample groups with a scenario-balanced distribution. Each task in the questionnaire asks, ``Which method generates more realistic and accurate images?'' with randomized option ordering to ensure unbiased results. As shown in Fig.~\ref{fig:user_study}, our method demonstrates significant superiority across all benchmarks.

\subsection{Failure Case Visualizations}
\begin{figure}
	\centering
	\includegraphics[width=\linewidth]{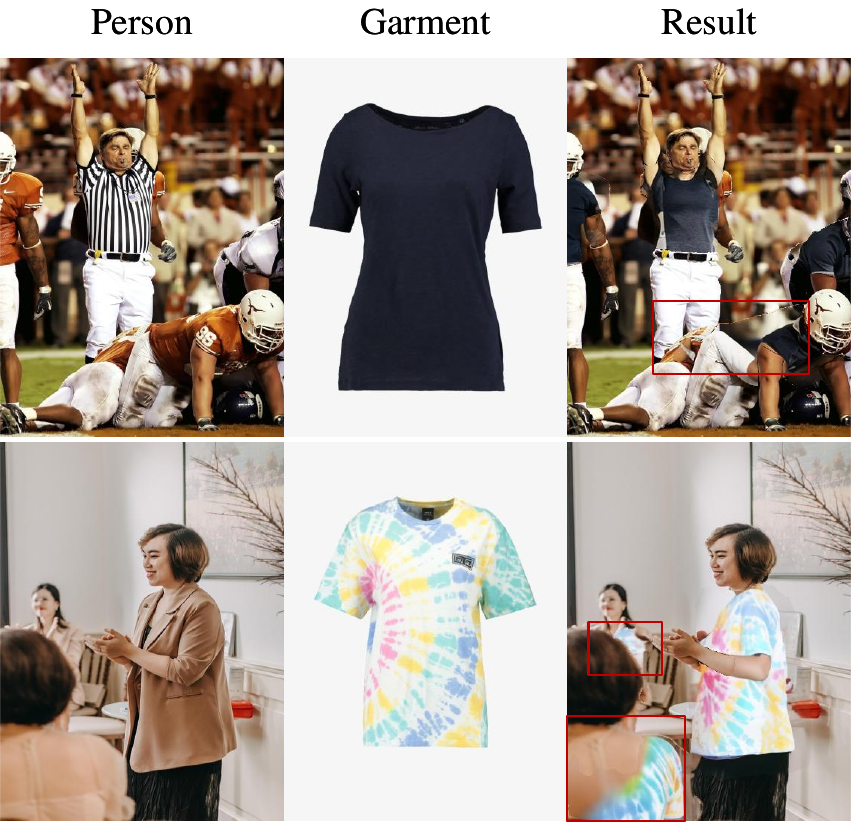}
	\captionof{figure}{Failure cases of our method.}
	\label{fig:limitation}
	\vspace{-2mm}
\end{figure}

We present several failure cases of OmniVTON in Fig.~\ref{fig:limitation}. As discussed in the main paper, our method encounters challenges in handling high-density crowds and targets with minimal visible body regions. These limitations primarily stem from OmniVTON's partial reliance on pre-trained modules such as OpenPose and TAPPS, whose predictions can be unreliable under such extreme conditions. Such observations point to a promising direction for future work towards more robust and adaptable universal virtual try-on systems.

\section{Additional Visual Results}
\subsection{Visual Comparisons with SOTAs}
\begin{figure*}
	\centering
	\includegraphics[width=\linewidth]{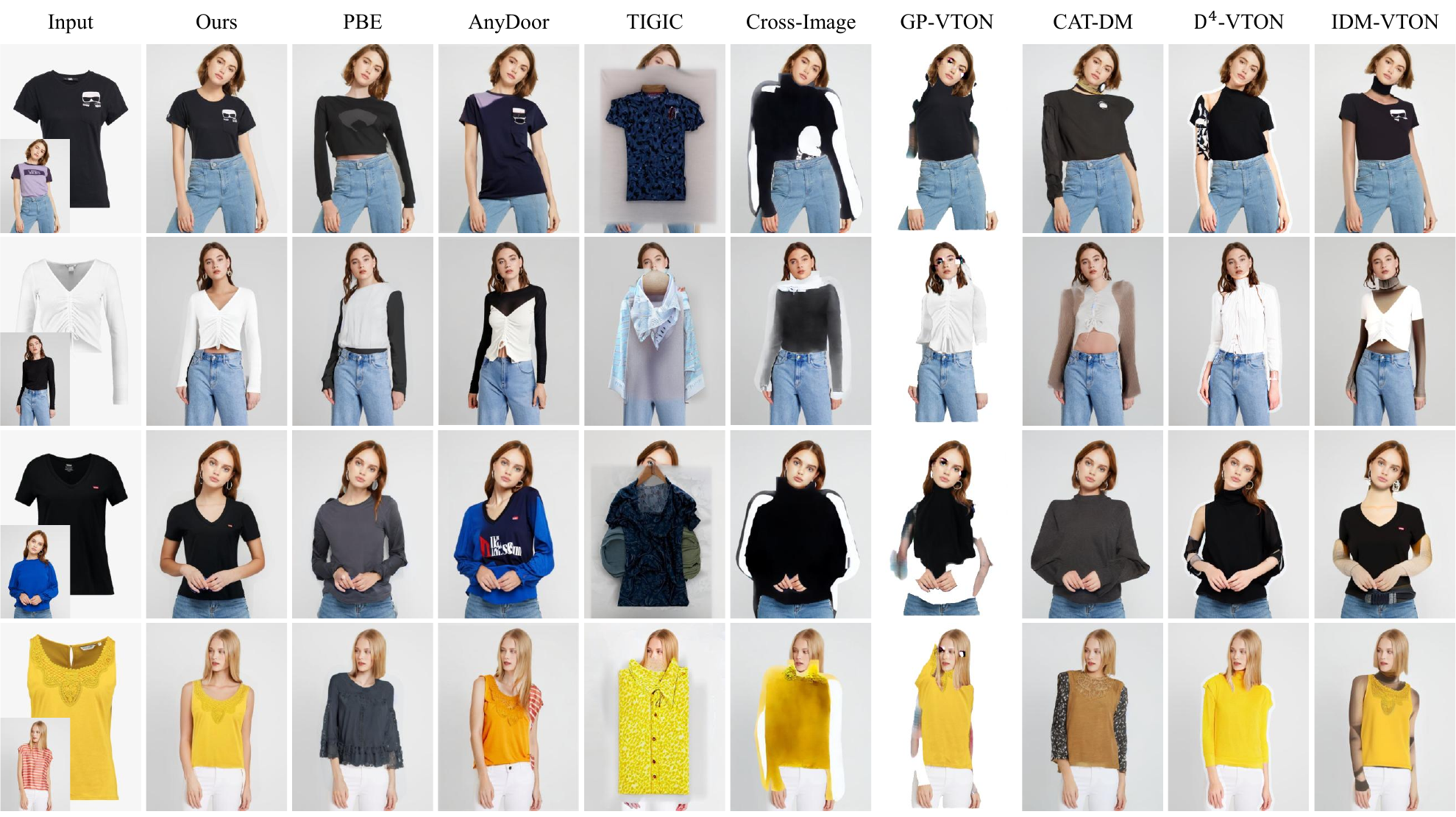}
	\vspace{-6mm}
	\captionof{figure}{Qualitative comparison on the VITON-HD dataset.}
	\label{fig:supp_vitonhd}	
\end{figure*}

\begin{figure*}
	\centering
	\includegraphics[width=\linewidth]{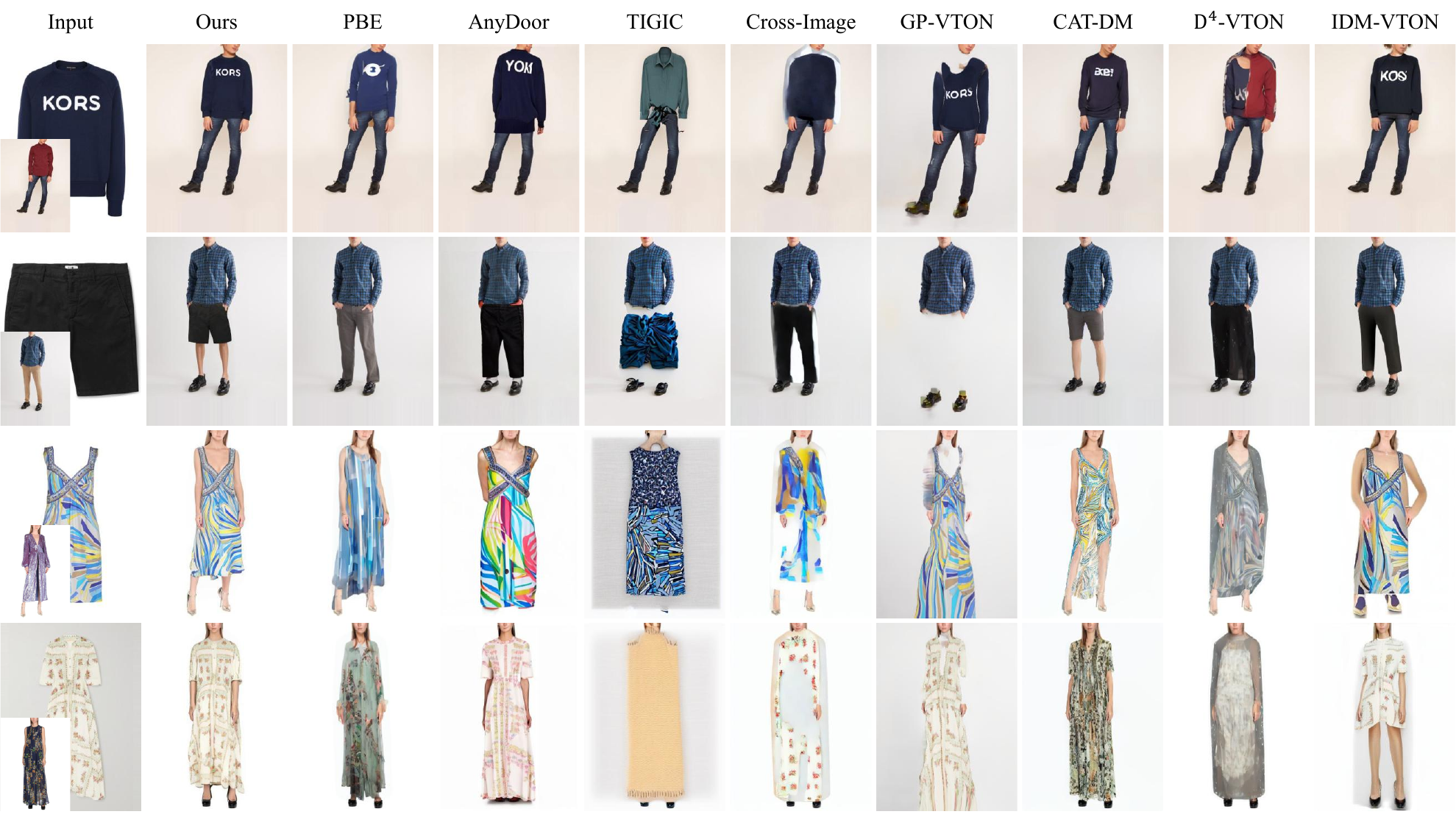}	
	\captionof{figure}{Qualitative comparison on the DressCode dataset.}
	\label{fig:supp_dresscode}	
\end{figure*}

\begin{figure*}
	\centering
	\includegraphics[width=\linewidth]{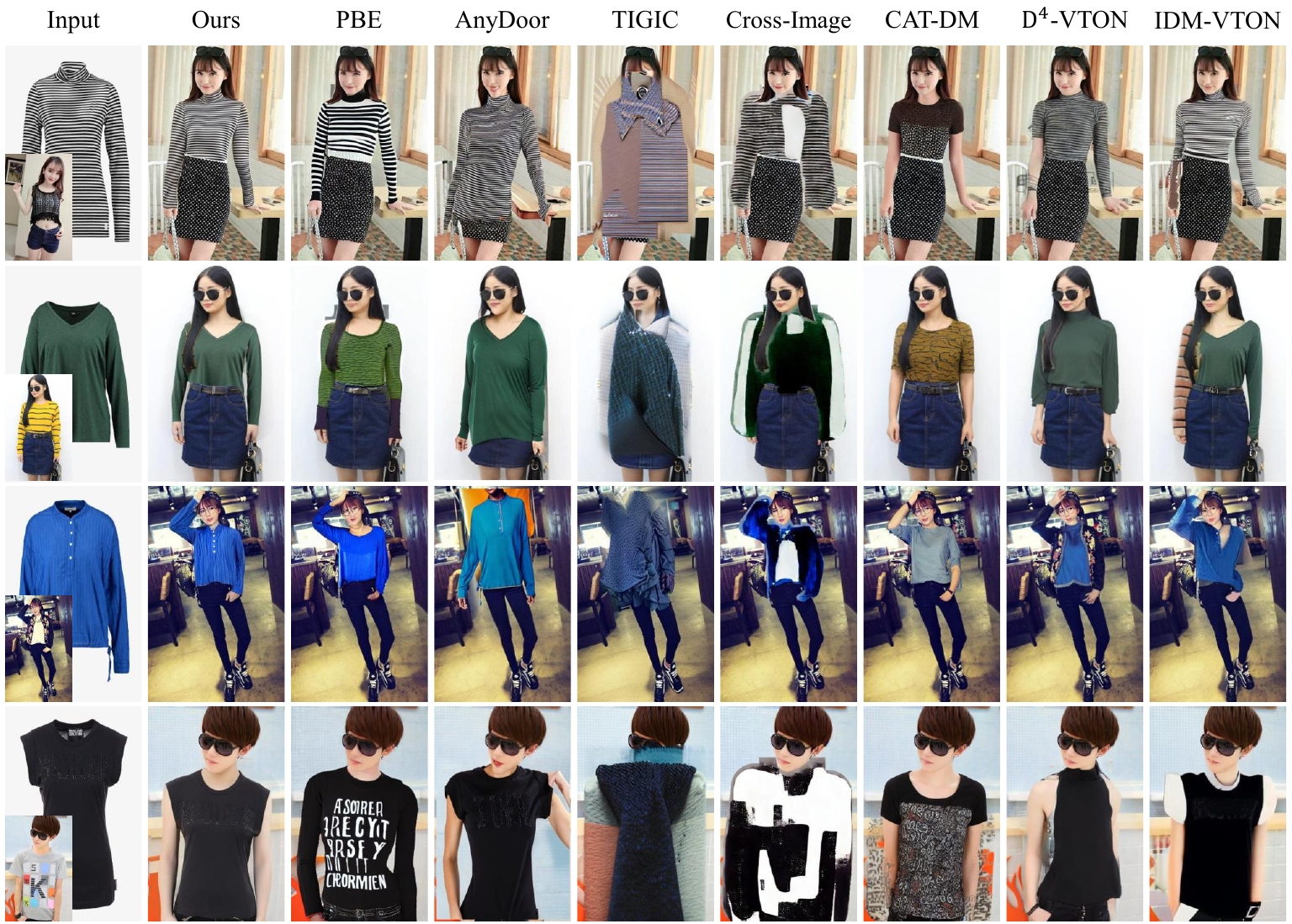}
	\vspace{-6mm}
	\captionof{figure}{Qualitative comparison for Shop-to-Street scenario on the StreetTryOn benchmark.}
	\label{fig:supp_street_sh2s}	
\end{figure*}

\begin{figure*}
	\centering
	\includegraphics[width=\linewidth]{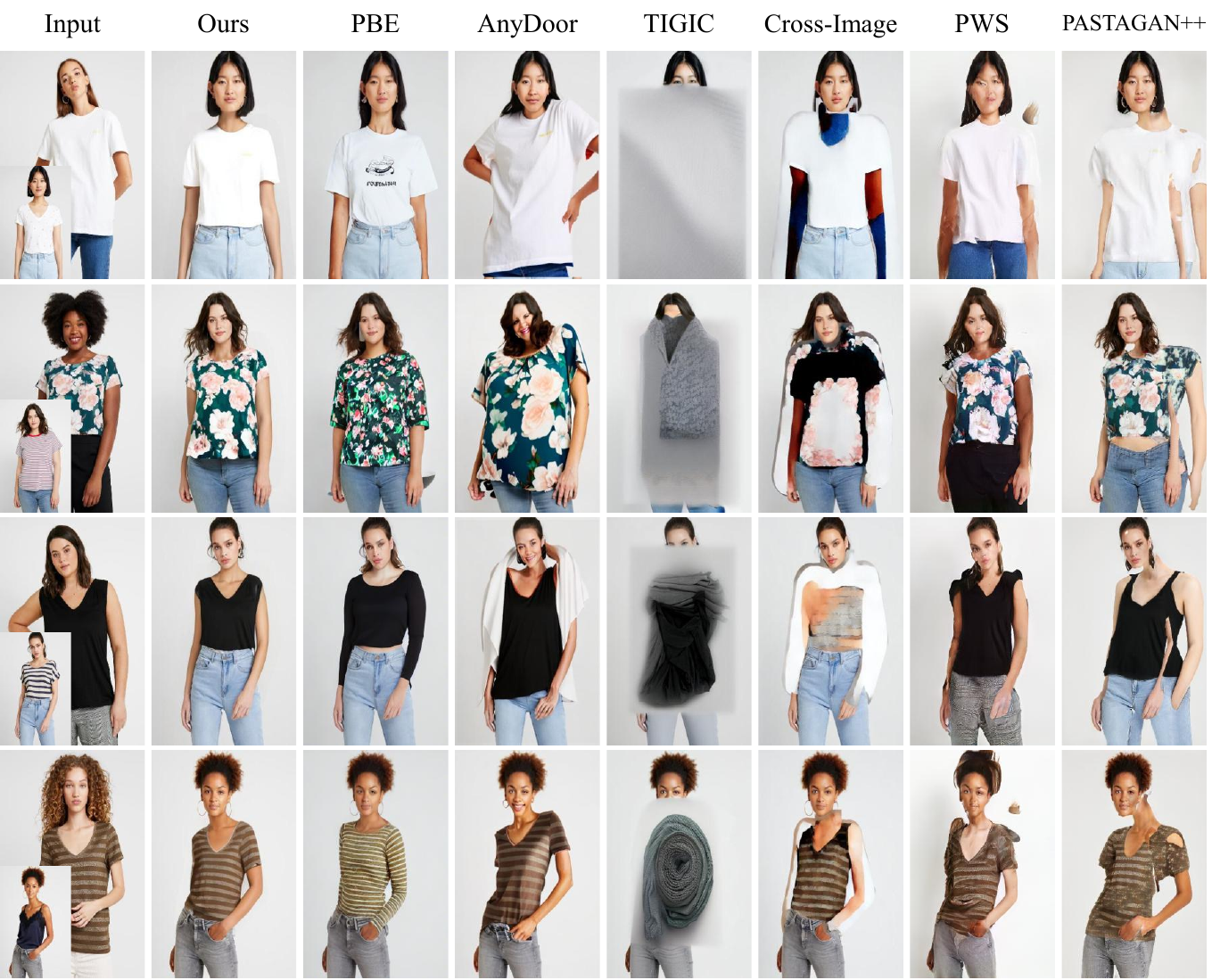}
	\vspace{-6mm}
	\captionof{figure}{Qualitative comparison for Model-to-Model scenario on the StreetTryOn benchmark.}
	\label{fig:supp_street_m2m}	
\end{figure*}

\begin{figure*}
	\centering
	\includegraphics[width=\linewidth]{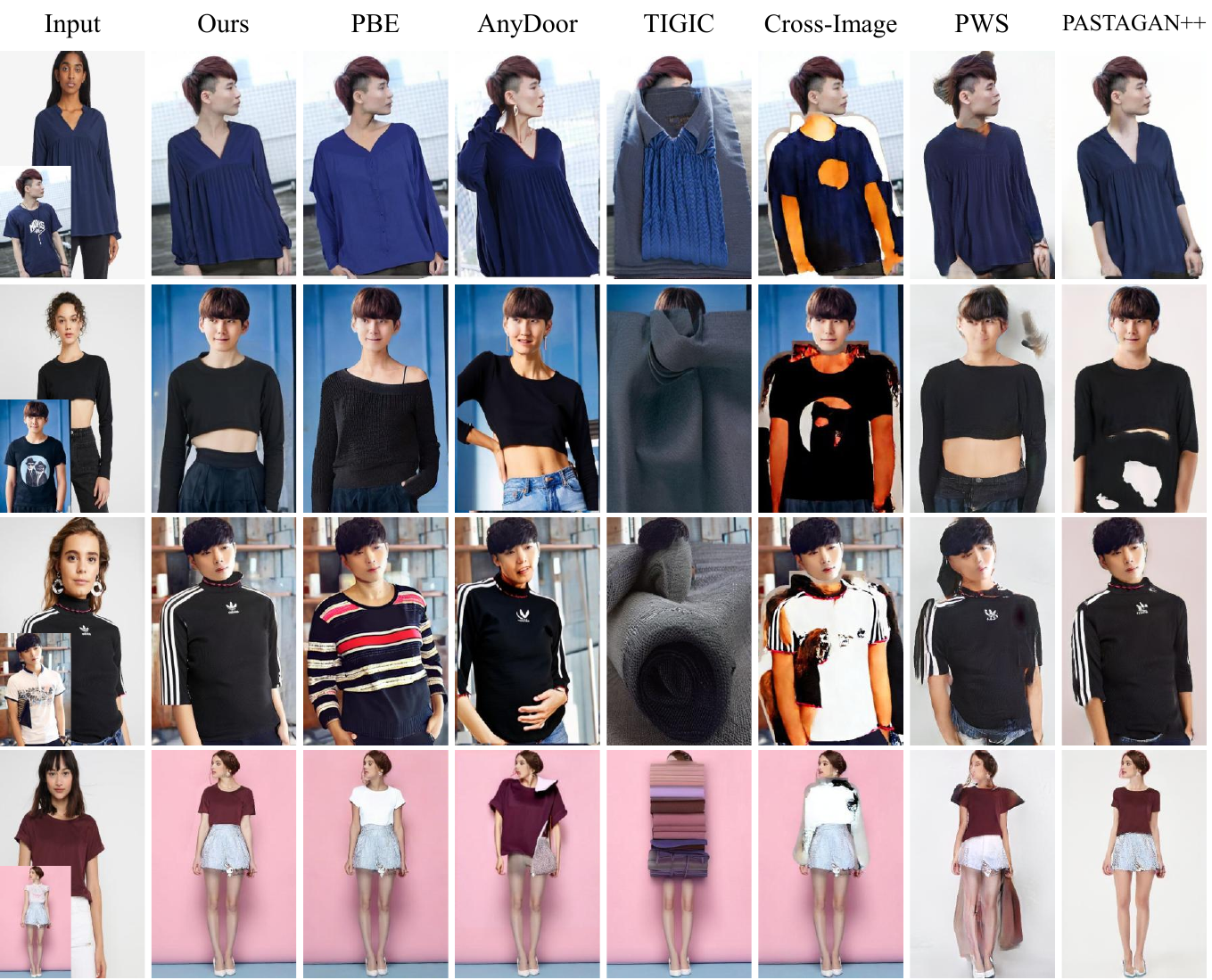}
	\vspace{-6mm}
	\captionof{figure}{Qualitative comparison for Model-to-Street scenario on the StreetTryOn benchmark.}
	\label{fig:supp_street_m2s}	
\end{figure*}

\begin{figure*}
	\centering
	\includegraphics[width=\linewidth]{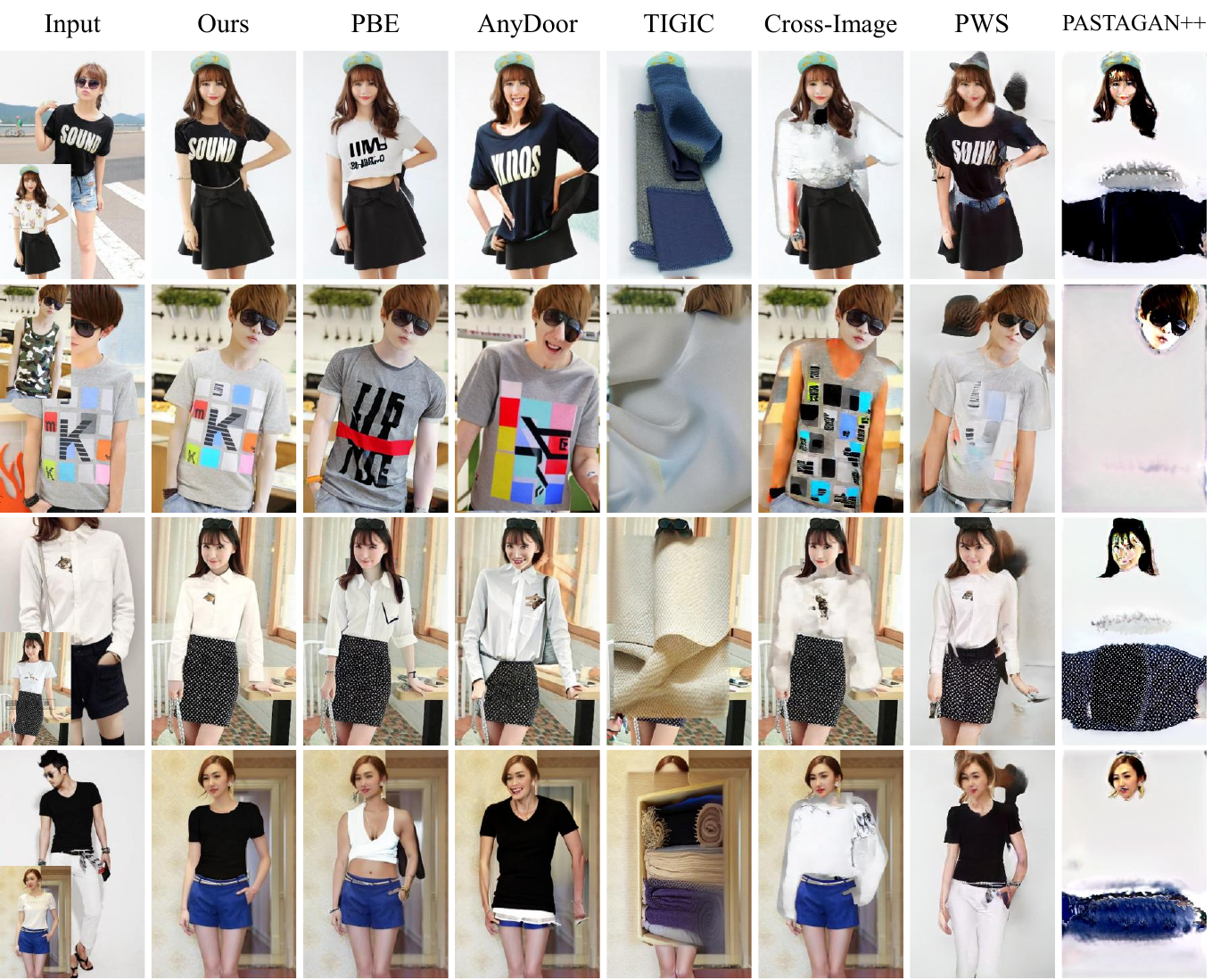}
	\vspace{-6mm}
	\captionof{figure}{Qualitative comparison for Street-to-Street scenario on the StreetTryOn benchmark.}
	\label{fig:supp_street_s2s}
\end{figure*}

Fig.~\ref{fig:supp_vitonhd} and Fig.~\ref{fig:supp_dresscode} present supplementary visual comparisons between OmniVTON and baseline methods on the VITON-HD and DressCode datasets, respectively. While Fig.~\ref{fig:supp_street_sh2s}, Fig.~\ref{fig:supp_street_m2m}, Fig.~\ref{fig:supp_street_m2s}, and Fig.~\ref{fig:supp_street_s2s} showcase detailed visualized results of different methods across four scenarios in the StreetTryOn benchmark.

\subsection{More Try-on Results}
\begin{figure*}
	\centering
	\includegraphics[height=0.8\textheight]{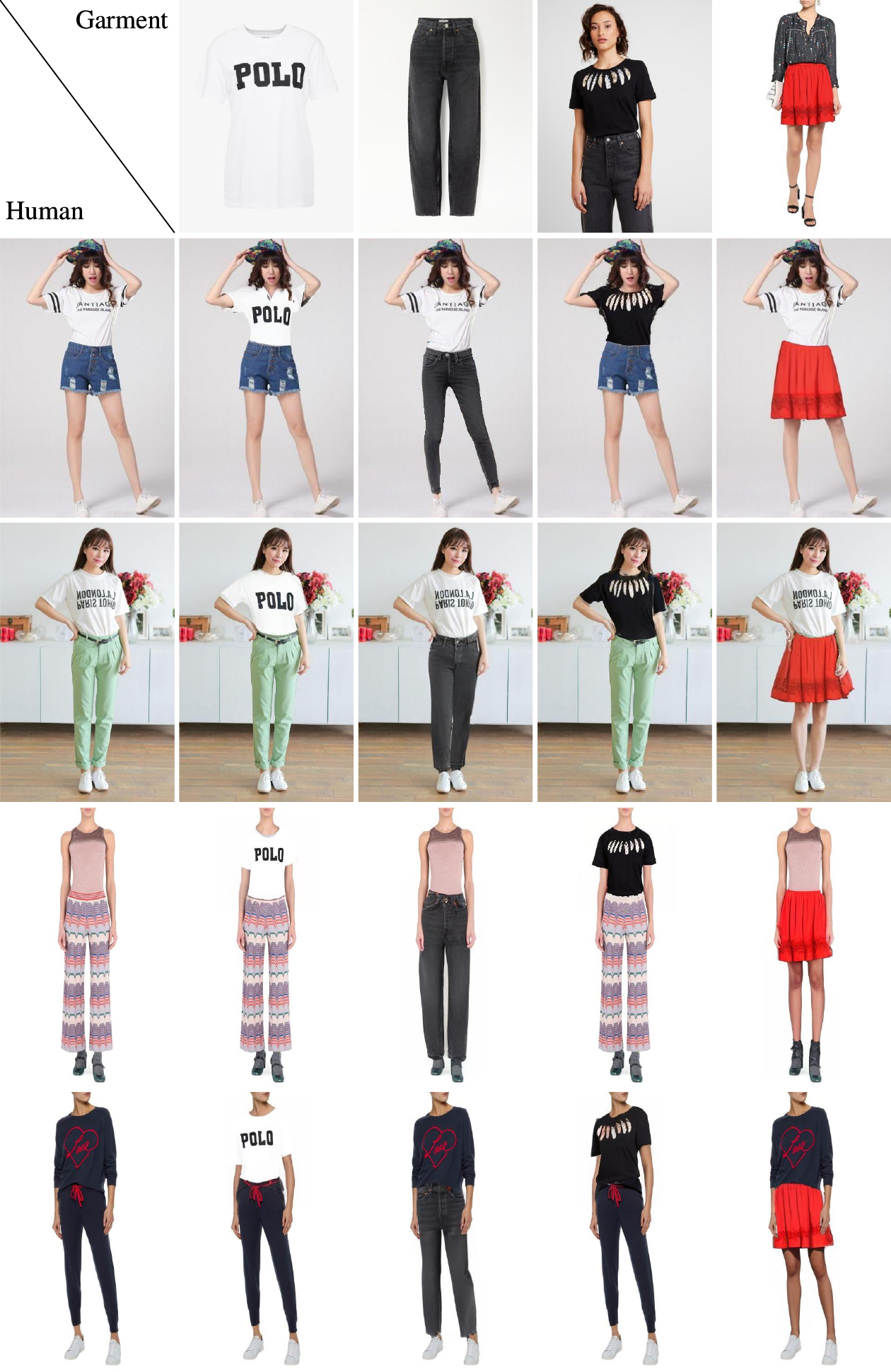}
	\captionof{figure}{More try-on results of OmniVTON across various clothing types and scenarios.}
	\label{fig:other_results}
	\vspace{-2mm}
\end{figure*}

As shown in Fig.~\ref{fig:other_results}, we further showcase various garment-model combinations, including virtual try-on results for lower-body garments and dresses under the Shop-to-Street scenario. This highlights OmniVTON's ability to overcome the technical barriers that previously limited the performance of StreetTryOn in this task.


\end{document}